%% file: PACE.tex
\documentclass{article}

\usepackage[preprint, nonatbib]{neurips_2024}

\usepackage[utf8]{inputenc} 
\usepackage[T1]{fontenc}    
\usepackage{url}            
\usepackage{booktabs}       
\usepackage{amsfonts}       
\usepackage{nicefrac}       
\usepackage{microtype}      
\usepackage{xcolor}         
\usepackage{colortbl}
\definecolor{tblcolor}{HTML}{BEF8D4}

\usepackage[pagebackref,breaklinks,colorlinks,bookmarks=false,urlcolor=black,citecolor={green!60!black}]{hyperref}

\input{preamble}

\title{PACE: Marrying generalization in PArameter-efficient fine-tuning with Consistency rEgularization}

\author{%
Yao Ni$^{\dagger}$ \quad Shan Zhang$^{\ddagger,\dagger}$ \quad Piotr Koniusz\thanks{The corresponding author. This paper is accepted by NeurIPS 2024 as spotlight.}$\;^{,\S,\dagger}$\\
$^{\dagger}$The Australian National University \quad $^\S$Data61$\!${\color{red}\heart}CSIRO \\
$^{\ddagger}$Australian Institute for Machine Learning, The University of Adelaide \\
{\tt\small $^{\dagger}$yao.ni@anu.edu.au $^{\ddagger}$shan.zhang@adelaide.edu.au $^\S$piotr.koniusz@data61.csiro.au}
}

\begin{document}

\maketitle

\begin{abstract}
  Parameter-Efficient Fine-Tuning (PEFT) effectively adapts pre-trained transformers to downstream tasks. However, the optimization of tasks performance often comes at the cost of generalizability in fine-tuned models. To address this issue, we theoretically connect smaller weight gradient norms during training and larger datasets to the improvements in model generalization. Motivated by this connection, we propose reducing gradient norms for enhanced generalization and aligning fine-tuned model with the pre-trained counterpart to retain knowledge from large-scale pre-training data. Yet, naive alignment does not guarantee gradient reduction and can potentially cause gradient explosion, complicating efforts to manage gradients. To address such an issue, we propose \our, marrying generalization of PArameter-efficient fine-tuning with Consistency rEgularization. We perturb features learned from the adapter with the multiplicative noise and ensure the fine-tuned model remains consistent for same sample under different perturbations. Theoretical analysis shows that \our not only implicitly regularizes gradients for enhanced generalization, but also implicitly aligns the fine-tuned and pre-trained models to retain knowledge. Experimental evidence supports our theories. PACE surpasses existing PEFT methods in visual adaptation tasks (VTAB-1k, FGVC, few-shot learning, domain adaptation) showcasing its potential for resource-efficient fine-tuning. It also improves LoRA in text classification (GLUE) and mathematical reasoning (GSM-8K). The code is available at \href{https://github.com/MaxwellYaoNi/PACE}{\color{red}{github.com/MaxwellYaoNi/PACE}}.
\end{abstract}

\section{Introduction}
Transformers \cite{transformer}, with the self-attention mechanism \cite{bahdanau2014neural}  capturing long-range dependencies in data, succeed in various deep learning tasks, including image classification (ViT \cite{vit}), multimodal learning (CLIP \cite{clip}), image synthesis (StableDiffusion \cite{rombach2022high}), semantic segmentation (SAM \cite{kirillov2023segment}) and text generation (LLaMA \cite{touvron2023llama}). The success of transformers can be largely attributed to the availability of abundant data, such as ImageNet \cite{imagenet} and Laion5B \cite{laion}, which empower researchers to scale up these models by training them under an enormous number of parameters.

Such huge models, with knowledge from large-scale pre-training \cite{su2021transferability}, constitute on foundation models that can be easily adapted to various downstream tasks through full fine-tuning or linear probing \cite{he2020momentum}, eliminating the need for task-specific model design \cite{chen2022adaptformer}. However, full fine-tuning is storage-intensive and infeasible for maintaining separate model weights as the number of tasks grows, while linear probing, which only trains the last head layer, yields inferior adaptation performance.

To overcome these limitations, Parameter-Efficient Fine-Tuning (PEFT) \cite{houlsby2019parameter} fine-tunes only a small subset of parameters, thereby reducing storage requirements while surpassing the performance of full fine-tuning and linear probing. These advantages have popularized PEFT and inspired the development of various PEFT methods for deep learning tasks, which can be categorized into two groups: those increasing inference cost and cost-efficient ones. The first group introduces additional learning branches, such as non-linear adapters \cite{adapter, chen2022adaptformer}, or concatenates learnable parameters with input tokens, \eg, visual prompts \cite{vpt, noah, oh2023blackvip}, increasing inference cost. The second group, focuses on cost-efficiency by lower-rank adaptation in linear layers \cite{glora, hu2022lora}, or affine transformations such as SSF \cite{Lian_2022_SSF} and RepAdapters \cite{repadapter}, which can be reparameterized during inference for efficiency.

Despite the superiority and efficiency of PEFT, prioritizing optimization for downstream tasks compromises the generalizability of fine-tuned models, yielding suboptimal performance. Although some analyses have been conducted on PEFT \cite{su2021transferability, hu2022sparse, fu2023effectiveness, wang2024universality, li2024adaptergnn}, they fail to fully explain the generalization of PEFT, leading to ineffective strategies for improving generalization. 

To address this gap in understanding generalization in PEFT, we establish a theoretical connection from generalization theory: smaller weight gradient norms and larger data volumes contribute to better generalization. Motivated by this, we propose reducing weight gradient norms and aligning output space of the fine-tuned model with the pre-trained one to retain knowledge captured from large pre-training data. Yet, theoretical analyses reveal this naive alignment does not guarantee gradient regularization and can even cause gradient explosion, complicating efforts for gradient management. To address this issue, we propose perturbing features learned from the adapter with multiplicative noise and constraining the network output to be consistent across different perturbations. 

Our method, called PACE, marries generalization of PArameter-efficient fine-tuning with Consistency rEgularization. Its name, PACE, reflects our goal of keeping the output behavior of the fine-tuned model \textit{in pace with} the pre-trained one. Despite its simplicity, theoretical analysis confirms that \our not only implicitly regularizes weight gradients for better generalization but also implicitly aligns the fine-tuned model with the pre-trained counterpart to retain knowledge from large-scale pre-training data. Experimental evidence supports our theories. \our improves existing PEFT methods, achieving superior results across six adaptation benchmarks. Our key contributions are:
\renewcommand{\labelenumi}{\roman{enumi}.}
\begin{enumerate}[leftmargin=0.7cm]
    \item We establish a theory connecting smaller weight gradient norms and larger datasets with enhanced generalization, motivating gradient reduction and model alignment for fine-tuning.\vspace{-0.1cm}
    \item We propose \our, a simple yet effective method perturbing features from adapters with multiplicative noise and constraining output of fine-tuned model to be consistent across perturbations.\vspace{-0.1cm}
    \item Our theoretical and empirical evidence confirms that \our implicitly regularizes gradients and aligns the fine-tuned model with the pre-trained one. \our excels on 4 visual adaptation tasks.\vspace{-0.1cm}
    \item We provide novel theoretical explanations of how gradient penalization and consistency regularization benefit generalization, offering fundamental insights applicable across deep learning. \vspace{-0.2cm}
\end{enumerate}

\section{Related work}
\textbf{Parameter-Efficient Fine-Tuning (PEFT).} LoRA \cite{hu2022lora} uses low-rank decomposition to reduce parameters and treats adapters as side paths. SSF \cite{Lian_2022_SSF} proposes affine transformations on latent features. FacT \cite{jie2023fact} decomposes and reassembles parameter matrices in ViT.  Surgical fine-tuning \cite{lee2023surgical} 
of 
different network parts improves adaptation to distribution shifts. FLoRA \cite{wen2023batched} performs a batched low-rank adaptation. GLoRA \cite{glora} unifies cost-efficient PEFT methods. NOAH \cite{noah} uses parameter search on neural prompts. ARC \cite{dong2024efficient} leverages cross-layer ViT similarity, parameter-sharing adapter and scaling factors for lower fine-tuning cost. RLRR \cite{dong2024low} incorporates a residual term for flexibility while preserving pre-trained representation. RepAdapter \cite{repadapter} reparameterizes adapters for efficient inference. Res-tuning \cite{jiang2024res} unbinds tuners from the backbone for memory efficiency. Zhao \etal \cite{zhao2024tuning} show impressive fine-tuning results by tuning layernorm in attention. OFT \cite{Qiu2023OFT} and BOFT \cite{liu2024boft} propose orthogonal fine-tuning to preserve hypersphere energy between neurons.

\textbf{Consistency Regularization.} Fixmatch \cite{sohn2020fixmatch} applies consistency regularization over augmented images for semi-supervised learning.
Openmatch \cite{saito2021openmatch} utilizes it on outlier predictions for open-set semi-supervised learning. R-Drop \cite{wu2021r} applies it to transformers \cite{transformer} with dropout for NLP tasks. CR \cite{zhang2019consistency} applies it over augmented real and fake images for GAN training. CAGAN \cite{ni2018cagan} enforces consistency on discriminators with dropout for GAN training. Despite the empirical success of consistency regularization demonstrated by previous works, theoretical analysis is lacking. While NICE \cite{ni2024nice} demonstrates that consistency regularization lowers latent feature gradients for stable GAN training, it fails to reveal reduced weight gradient for enhanced generalization. Our study goes beyond prior works by providing a theoretical link between smaller weight gradients and improved generalization, effectively marrying generalization of PEFT with consistency regularization.

 \textbf{Generalization of Fine-Tuning.} Li \etal \cite{li2021improved} constrain the fine-tuned model's closeness to the pre-trained model in weight space. Fu \etal \cite{fu2023effectiveness} induce sparsity on PEFT for better generalization. Wang \etal \cite{wang2024universality} studies  generalization of PEFT fine-tuning graph neural network. Zhang \etal \cite{zhang2024gbboost} employ rank-1 gradient boosting (GB) updates supported by the GB theoretical framework.  
 VioLET \cite{wang2023violet}, PromptSRC \cite{khattak2023self} and CoPrompt \cite{roy2024consistencyguided} naively align the fine-tuned model with the pre-trained one for enhanced generalization or avoiding forgetting. Additionally, L2SP \cite{xuhong2018explicit}, DELTA \cite{li2019delta}, and FTP \cite{tian2024fast} aim to retain pre-trained knowledge by aligning fine-tuned models with pre-trained ones, reducing distance in weight space, feature space and using projected gradient descent, respectively. 
 However, they fail to provide a theoretical analysis for this alignment. Our study goes beyond understanding generalization of PEFT by discovering the benefits of gradient regularization and model alignment. We propose PACE to match both requirements, paving a comprehensive understanding for PEFT. 

\textbf{Gradient regularization.} Previous studies have empirically shown that gradient regularization improves  performance \cite{varga2017gradient, zhao2022penalizing, Ni_2024_CVPR, ni2022manifold} and adversarially robust accuracy \cite{junhao_vlms}. However, they lack  theoretical connection between smaller gradient norms and better generalization \cite{foret2021sharpnessaware, zhang2024semantic, cha2021swad}. We  bridge this gap by establishing a fundamental theory between reduced gradient norms and improved generalization, providing a solid foundation for future research on enhancing generalization.

\section{Approach}
We begin with a unified perspective on cost-efficient PEFT based on GLoRA \cite{glora}, linking generalization with gradients and large-scale data, and motivating the alignment of the fine-tuned model with the pre-trained model to leverage its knowledge. We identify limitations of naive alignment in gradient regularization and introduce PACE, which implicitly enhances gradient regularization and model alignment. We conclude with theoretical justification and efficient implementations.

\subsection{A unified perspective on cost-efficient PEFT methods}
The transformer architectures \cite{transformer, vit} have excelled in natural language processing and computer vision tasks through their powerful sequential modeling capabilities. This success stems from their ability to process text/image tokens through $L$ transformer blocks, where each block contains self-attention and MLP modules primarily composed of linear layers. These linear layers enable the self-attention mechanism to capture long-range dependencies, allowing transformers to achieve superior performance when scaled to a huge  number of parameters and trained on extensive datasets.

With massive parameters, pre-trained on large-scale data, transformers serve as foundation models that can be fine-tuned for downstream tasks using limited data. However, fully fine-tuning all parameters for various downstream tasks requires substantial memory and can lead the forgetting of pre-trained knowledge. To alleviate this without increasing inference cost, adapters with lightweight parameters are often preferred for fine-tuning. Let $\bar{h}_0(\cdot)$ be a transformation within the pre-trained transformer. Current adapters can be unified as introducing a residual branch $\Delta\bar{h}$ to form a new transformation $\bar{h}$:
\begin{equation}
    \bar{h}(\va) = \bar{h}_0(\va)+\Delta\bar{h}(\va).
\end{equation}
Here,  $\va$ is the input and $\bar{h}_0(\cdot)$ can represent MLP modules, as in Adapter \cite{adapter} and AdaptFormer \cite{chen2022adaptformer}, or linear layers in self-attention and MLP modules, as in \cite{hu2022lora, glora, dettmers2024qlora, kopiczko2024vera}. In SSF \cite{Lian_2022_SSF}, $\bar{h}_0(\cdot)$ is the identity mapping and $\Delta\bar{h}(a)=\va\odot(\vgamma-\vone)+\vbeta$ with $\vgamma$ and $\vbeta$ as affine transformation parameters.

Given that linear layers are key components in transformer, tuning them offers a flexible and effective way to adapt models to downstream tasks. This work focuses on methods that tune the linear layer without increasing inference cost. Let $(\mW_0,\vb_0)$, $(\Delta\mW, \Delta\vb)$, and $(\mW, \vb)$ be the parameters of pre-trained model, adapter and fine-tuned model, respectively, where $\mW_0, \Delta\mW, \mW\in\bbR^{\dout\times\din}$ and $\vb_0,\Delta\vb, \vb\in\bbR^{\dout}$. Fine-tuning a linear layer in self-attention or MLP module can be formed as:
\begin{align}
    h(\va)&= \mW\va+\vb=(\mW_0+\Delta\mW)\va+(\vb_0+\Delta\vb)\nonumber\\
          &= h_0(\va)+\Delta h(\va)=(\mW_0\va+\vb_0)+(\Delta\mW\va+\Delta\vb).
\end{align}
Based on GLoRA \cite{glora}, cost-efficient PEFT methods for linear layers vary in the form of $\Delta\mW, \Delta\vb$:

\noindent\textbf{\loraadd}: $\Delta\mW=\Wdown\Wup, \Delta\vb=\vblora$ where $\Wdown\in\bbR^{\dout\times r}, \Wup\in\bbR^{r\times \din}$, and $r$ is the rank.

\noindent\textbf{\loramul}: $\Delta\!\mW\!=\!\mW_{\!0}\!\odot\!(\Wdown\Wup)$, $\Delta\vb\!=\!\vb_0\!\odot\!\vblora$, including RepAdapter \cite{repadapter} via reparameterization.

\noindent\textbf{\vptadd}: $\Delta\mW$ is zero, $\Delta\vb=\mW_0\mP$, with learnable $\mP\in\bbR^{\din\times1}$ as layer-wise visual prompt. We use \vptadd\  to differentiate from VPT \cite{vpt}, which concatenates $\mP$ with tokens, increasing inference cost.

\subsection{Generalization of deep neural networks}
Having established a unified perspective on cost-efficient PEFT, we now motivate our method from a perspective on improving generalization of neural networks to enhance performance on unseen data. Consider a network $f:=\phi(g(x))$ with $l$ layers, where $g$ is feature extractor and $\phi$ is the classification head. Let $\vtheta:=\{(\mW^{(i)}, \vb^{(i)})\}_{i=1}^l$ be the parameter set with dimension $d$ and $\calDn:=\{(\vxi, \vyi)\}_{i=1}^n$ be the training set of size $n$ drawn {\em \iid} from distribution $\scrD$, which contains infinite data. The following lemma from \cite{foret2021sharpnessaware} explains  the relationship between the empirical and population loss. 
\begin{lemma}\label{lemma:sam}
    (Theorem 1 from \cite{foret2021sharpnessaware}) Let $\calL_{\calDn}(\vtheta)$ be the empirical loss function over $f$ on training set $\calDn$ and $\calL_{\scrD}(\vtheta)$ be the population loss. For any $\rho>0$, with high probability over $\calDn\sim\scrD$, we have
    \vspace{-0.1cm}%
    \begin{equation}
    \vspace{-0.1cm}
        \calL_{\scrD}(\vtheta) \leq \max_{\lVert\vepsilon\rVert_2\leq\rho}\calL_{\calDn}(\vtheta+\vepsilon)+R\Big(\frac{\lVert\vtheta\rVert_2^2}{\rho^2}, \frac{1}{n}\Big),\label{eq:sam}
    \end{equation}
    where $R:(\bbR_+, \bbR_+)\rightarrow\bbR_+$ is an increasing function (under conditions on $\calL_\scrD(\vtheta)$ and $n$ as in \S \ref{sup:subsec:sample_size}).
\end{lemma}
Lemma \ref{lemma:sam} bounds the population loss by the empirical loss with perturbed weights, indicating that a minimal empirical loss increase from small weight perturbations implies low population loss.

By observing that the maximum of $\calL_{\calDn}$ is achieved at $\vepsilon=\frac{\rho\nablatheta}{\lVert\nablatheta\rVert_2}$, where $\nablatheta$ is the gradient of $\calL_{\calDn}$ at $\vtheta$,  and performing a Taylor expansion of $\calL_{\calDn}$ around $\vtheta$,  we formulate the following theorem.

\begin{theorem}
\label{thm:grad}
Denote $\nablatheta$ as the gradient and $\lambdaHmax$ as the largest eigenvalue of the Hessian matrix $\Htheta$ of $\calL_{\calDn}$ at $\vtheta$. For any $\rho>0$, with high probability over training set $\calDn\sim\scrD$, we have
\vspace{-0.1cm}%
\begin{equation}
\vspace{-0.1cm}%
     \calL_{\scrD}(\vtheta)\leq\calL_{\calDn}(\vtheta)+\rho\lVert\nablatheta\rVert_2+\frac{\rho^2}{2}\lambdaHmax+R\Big(\frac{\lVert\vtheta\rVert_2^2}{\rho^2}, \frac{1}{n}\Big).
\end{equation}
Here, higher-order terms from the Taylor expansion are incorporated into $R\Big(\frac{\lVert\vtheta\rVert_2^2}{\rho^2}, \frac{1}{n}\Big)$, which is related to weights norm and inversely related to the training data size $n$.
\end{theorem}

Theorem \ref{thm:grad} (proof in \S \ref{sup:sec:proof_thm_grad}) outlines strategies for enhancing generalization. They involve regularizing weight norms and the largest Hessian eigenvalues, and crucially, increasing data size $n$ and reducing the weight gradient norms (illustrated in Figure \ref{fig:theorems_intuitive}). However, excessive reduction should be avoided as it could impair network's representation capacity, yielding higher empirical and population loss.

\subsection{Motivation and limitation of aligning the fine-tuned model with the pre-trained model}
Theorem \ref{thm:grad} emphasizes that large-scale data and smaller gradient magnitudes are essential for better generalization in neural network training. Therefore, aligning the fine-tuned model with the pre-trained one is crucial, as it ensures retention of knowledge obtained  from large-scale data, preserving generalization. PEFT methods, often outperforming full fine-tuning, achieve this alignment by limiting the number of trainable parameters, restricting the model's capacity to deviate from the pre-trained one. However, the training objective prioritizes downstream task performance, compromising alignment with pre-trained knowledge. While sparsity regularization \cite{fu2023effectiveness} and weight decay on adapter weights help, they do not ensure alignment, as even small weight changes can lead to significant divergence in output space \cite{wu2020adversarial, he2020defending, foret2021sharpnessaware}. Therefore, we propose to achieve the alignment by reducing the FP-distance (output distance between fine-tuned and pre-trained models on training samples):
{{\setlength{\abovedisplayskip}{0.1cm}%
\setlength{\belowdisplayskip}{0.1cm}%
\begin{equation}
    \Dfp(\vtheta) = \frac{1}{n}\sum_{i=1}^n\lVert f(\vxi; \vtheta) - f(\vxi; \vtheta_0)\rVert_2^2, \quad \vtheta=\vtheta_0+\Delta\vtheta,\label{eq:fp_distance}
\end{equation}}%
where $\vtheta, \vtheta_0, \Delta\vtheta\in\bbR^d$ are parameters for the fine-tuned model, pre-trained model and the adapter.

While reducing FP-distance keeps the fine-tuned model close to the pre-trained model, thus preserving its knowledge, it does not ensure reduced gradient magnitudes, leading to suboptimal generalization. To understand the gradient-related limitations in this alignment, we assume $\Delta\vtheta$ is small enough for a Taylor expansion approximation. Following standard practices \cite{foret2021sharpnessaware, zhang2021how, denoising_JMLR}, we perform the expansion up to the second-order terms. Given the independence between elements in squared $L_2$ distances (\S\ref{sup:subsec:onedim}) and to simplify our theories, we analyze a one-dimensional output for a single {\em\iid} sample, which leads us to the following proposition.
\begin{prop}\label{prop:fp}
    Assuming $\Delta\vtheta$ is small, denote $f(\vtheta)\in\bbR$ as the one-dimensional output for $\vx$, with $\vnabla$ and $\mH$ as its gradient and Hessian at $\vtheta$. FP-distance over $\vx$ can be decomposed as follows:
    {\setlength{\abovedisplayskip}{0.2cm}%
\setlength{\belowdisplayskip}{0.cm}
    \begin{align}
        [f(\vtheta)-f(\vtheta_0)]^2&=[f(\vtheta)-f(\vtheta-\Delta\vtheta)]^2 \approx\big[f(\vtheta)-[f(\vtheta)-\Delta\vtheta^T\vnabla+\frac{1}{2}\Delta\vtheta^T\!\mH\Delta\vtheta]\big]^2\nonumber\\
                       &\approx[\Delta\vtheta^T\vnabla-\frac{1}{2}\Delta\vtheta^T\mH\Delta\vtheta]^2.\label{eq:fp_taylor}
    \end{align}}
\end{prop}%
\vspace{-0.1cm}%
Prop. \ref{prop:fp} establishes the relationship between weight gradients, adapter weights, and FP-distance. However, it remains unclear if it regulates gradients. Our experiments show that minimizing FP-distance can sometimes increase gradient magnitude, complicating efforts for managing gradient.

\begin{figure}[t]
    \centering
    \includegraphics[width=\linewidth]{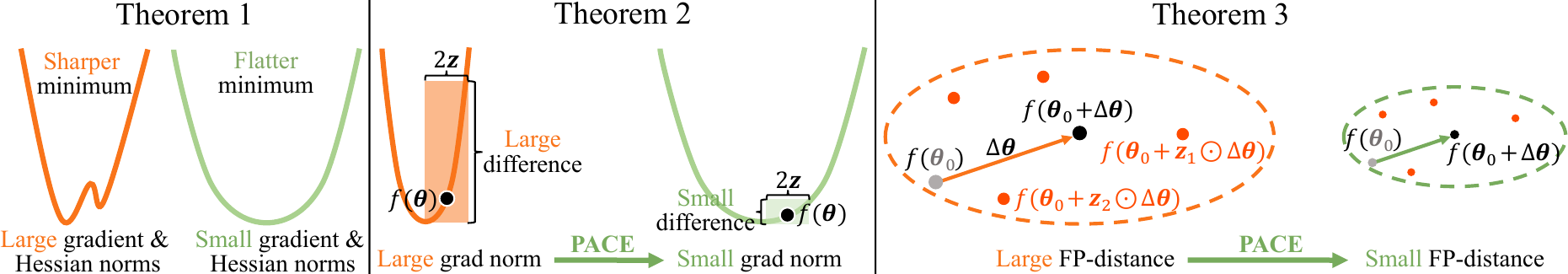}
    \caption{Thm.~\ref{thm:grad}: A flatter minimum has smaller gradient and Hessian norms, yielding better generalization. Thm. \ref{thm:pace}: Large gradient norms indicate large differences among perturbations. PACE minimizes these differences, reducing gradient norms. Thm. \ref{thm:pace_compare}: Minimizing all pairs of distances between $f(\vtheta_0\!+\!\vz_1\!\odot\!\Delta\vtheta)$ and $f(\vtheta_0\!+\!\vz_2\!\odot\!\Delta\vtheta)$ where $\vz_1, \vz_2\!\sim\!\calN(\vone, \sigma^2\mI)$ also reduces FP-distance (between fine-tuned $f(\vtheta_0\!+\!\Delta\vtheta)$ and pre-trained $f(\vtheta_0)$), especially when $\vz_1\!\!=\!\!\vone$, $\vz_2 \!=\!\vzero$ or vice versa.} 
    \label{fig:theorems_intuitive}
    \vspace{-0.3cm}
\end{figure}

\subsection{Consistency regularization}
To achieve better generalization by both regularizing gradients and aligning the fine-tuned model with the pre-trined model, we propose a consistency regularization loss for $f$, encouraging invariance of $f$ to the same input under varying multiplicative noise perturbations on the adapter weights, as follows:%
\vspace{-0.1cm}%
\begin{equation}
\vspace{-0.1cm}
    \Dpace(\vtheta)=\frac{1}{n}\sum_{i=1}^n\bbE_{\vz_1,\vz_2}\lVert f(\vx_i;\vtheta_0+\vz_1\odot\Delta\vtheta)-f(\vx_i;\vtheta_0+\vz_2\odot\Delta\vtheta)\rVert_2^2,
\end{equation}
where $\vz_1,\vz_2\sim\calN(\vone, \sigma^2\mI)$ is the multiplicative noise applied on adapter weight. To understand the generalization benefits in this consistency regularization, we simplify the analysis by focusing on one-dimensional output for a single sample, resulting in the following theorem.
\begin{theorem}\label{thm:pace}
    Using notations from Prop. \ref{prop:fp}, let $f(\vtheta_0+\vz\odot\Delta\vtheta)\in\bbR$ be the one-dimensional output for $\vx$. Define $\Delta\theta_j$ as $j$-th element in $\Delta\vtheta$, $\nabla_j$ as the $j$-th element in $\vnabla$ and $H_{jk}$ as the $(j,k)$-entry in $\mH$. With $\vz_1,\vz_2\sim\calN(\vone, \sigma^2\mI)$, the consistency loss over $\vx$ can be approximated as:
    {\setlength{\abovedisplayskip}{0.2cm}%
    \setlength{\belowdisplayskip}{0.1cm}%
    \begin{align}
        &\bbE_{\vz_1, \vz_2}[f(\vtheta_0+\vz_1\odot\Delta\vtheta)-f(\vtheta_0+\vz_2\odot\Delta\vtheta)]^2\nonumber\\
        \approx&\colorbox{white}{$\!2\sigma^2\!\sum_j\Delta\theta_j^2\nabla_j^2\!+\!\sigma^4\!\sum_{j,k}\!\Delta\theta_k^2\Delta\theta_j^2H_{jk}^2$} = 2\sigma^2\lVert\Delta\vtheta\odot\vnabla\lVert_2^2+\sigma^4\lVert(\Delta\vtheta\Delta\vtheta^T)\odot\mH\rVert_F^2.\label{eq:pace_taylor}
    \end{align}}
\end{theorem}
Theorem \ref{thm:pace} (proof in \S \ref{sup:sec:proof_thm_pace}) shows that the consistency regularization essentially penalizes the first- and second-order gradients of $f$ at $\vtheta$ (illustrated in Figure \ref{fig:theorems_intuitive}), with the regularization strength controlled by the noise variance $\sigma^2$ and adaptively influenced by the magnitude of elements in adapter weight $\Delta\vtheta$. Thus, minimizing the consistency loss implicitly regularizes the gradients, improving generalization.

With the FP-distance in Prop. \ref{prop:fp} and consistency loss in Theorem \ref{thm:pace}, we establish their relationship as:
\begin{theorem}\label{thm:pace_compare}
With $d$ as the dimension of $\vtheta$, Eq. \ref{eq:fp_taylor} can be upper-bounded as:
{\setlength{\abovedisplayskip}{0.1cm}%
    \setlength{\belowdisplayskip}{-0.1cm}%
    \begin{align}
        [\Delta\vtheta^T\vnabla-\frac{1}{2}\Delta\vtheta^T\mH\Delta\vtheta]^2\leq2d\lVert\Delta\vtheta\odot\vnabla\rVert_2^2+d^2\lVert(\Delta\vtheta\Delta\vtheta^T)\odot\mH\rVert_F^2.
    \end{align}}
\end{theorem}
Theorem \ref{thm:pace_compare} (proof in \ref{sup:sec:proof_pace_compare}) establishes the relationship between Eq. \ref{eq:fp_taylor} and Eq. \ref{eq:pace_taylor}, showing Eq. \ref{eq:fp_taylor} is upper-bounded by terms involving $\lVert\Delta\vtheta\!\odot\!\vnabla\rVert_2^2$ and $\lVert(\Delta\vtheta\Delta\vtheta^T)\!\odot\!\mH\rVert_F^2$ which appear in Eq. \ref{eq:pace_taylor}. Reducing these terms results in a decrease in Eq. \ref{eq:fp_taylor}. Thus minimizing the consistency loss implicitly aligns the fine-tuned and pre-trained models (illustrated in Figure \ref{fig:theorems_intuitive}),  preserving pre-trained knowledge. 

\subsection{Efficient implementation of PACE}
\begin{figure}[t]
    \centering
    \includegraphics[width=\linewidth]{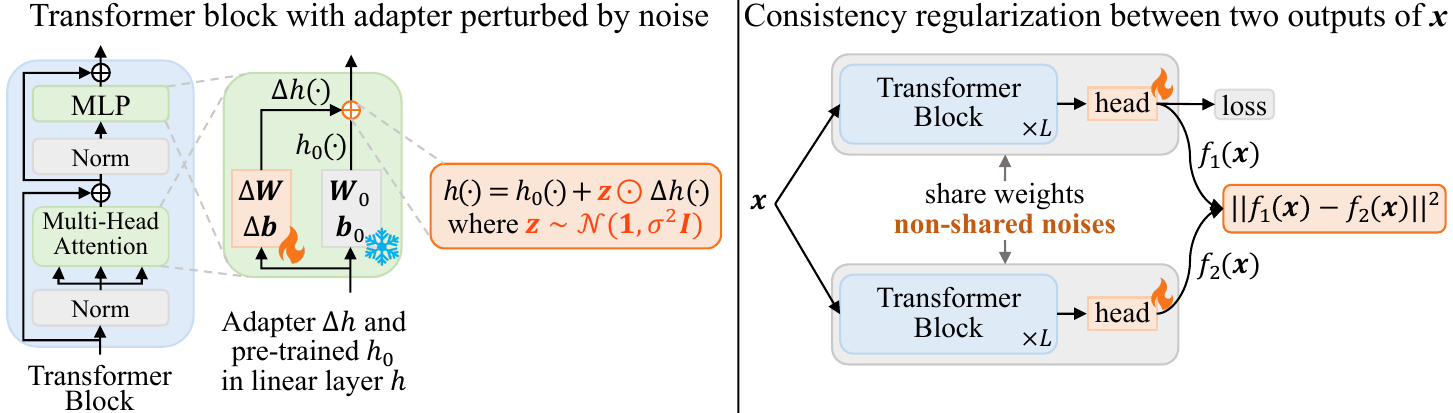}
    \caption{Our pipeline. Adapter $\Delta h(\cdot)$ and $h_0(\cdot)$ from pre-trained model form the linear layer $h$ of Multi-Head Attention and MLP in fine-tuned model. We perturb $\Delta h(\cdot)$ with multiplicative noise and ensure the network remains consistent to same inputs under varying perturbations.}
    \label{fig:pipeline}
    \vspace{-0.2cm}
\end{figure}

Providing different weight perturbations for each input in a mini-batch increases memory and computational demands. To avoid this, we perturb feature outputs from the adapter $\Delta h(\cdot)$, effectively simulating perturbation that shares noise across each row in the weight matrix $\Delta\mW$. Our simple pipeline is shown in Figure \ref{fig:pipeline}. Consider $\mX\in\bbR^{B\times T\times \din}$ as a batch of data where $B$ and $T$ are the batch and token sizes. The calculation for the linear layer of the fine-tuned model, which utilizes pre-trained weights $\mW_0,\vb_0$ and adapter weights $\Delta\mW, \Delta\vb$, processes an output size of $\dout$ as:
\begin{align}
    h_0(\mX)&=\mW_0\mX+\vb_0;\quad \Delta h(\mX)=\Delta\mW\mX+\Delta\vb,\\
    h(\mX) &= h_0(\mX)+\mZ\odot\Delta h(\mX).\label{eq:pace_noise}
\end{align}
Operator $\odot$ is the element-wise multiplication after expanding the left matrix $\mZ\in\bbR^{B\times\dout}\sim\calN(\vone, \sigma^2\mI)$ into $B\times T\times \dout$ where tokens within the same example share the same noise. Motivated by \cite{li2023dropkey}, the $\sigma$ decreases linearly as block depth increases. Let $f_1(\cdot)$ and $f_2(\cdot)$ be two networks share same weights but do not share the noise patterns. The loss function for \our is:
{\setlength{\abovedisplayskip}{0.1cm}%
\setlength{\belowdisplayskip}{0.1cm}%
\begin{align}
    \calL^\text{PACE} = \frac{1}{n}\sum_{i=1}^n\ell(f_1(\vxi), \vyi)+\lambda\lVert f_1(\vx_i)-f_2(\vx_i)\lVert_2^2,
\end{align}}%
where $\ell$ is the classification loss and $\lambda$ is a hyperparameter controlling regularization strength. During inference, noise and regularization are ommitted, $\Delta\mW, \Delta\vb$ are integrated with $\mW_0, \vb_0$ for efficiency:
{\setlength{\abovedisplayskip}{0.2cm}%
\setlength{\belowdisplayskip}{0.1cm}%
\begin{align}
    \mW =\mW_0+\Delta\mW; \quad \vb=\vb_0+\Delta\vb; \quad h(\mX) = \mW\mX+\vb.
\end{align}}

\textbf{Efficient \our variants.} In \S\ref{sup:sec:efficient-pace}, we present two variants that match the computational/memory costs of the baseline while achieving superior performance with substantially reduced resources.

\begin{table}[t!]
\begin{center}
\fontsize{8}{9}\selectfont
\setlength{\tabcolsep}{0.06cm}
\caption{Results on VTAB-1K with ViT-B/16. Mean Acc. is the average of group mean values.}
\label{tab:vtab-comparison}
\begin{tabular}{!{\vrule width \boldlinewidth}l|ccccccc|cccc|cccccccc|c!{\vrule width \boldlinewidth}}
\topline
\multirow{2}{*}{Method} & \multicolumn{7}{c|}{Natural} & \multicolumn{4}{c|}{Specialized} & \multicolumn{8}{c|}{Structured}&\\
\cline{2-20}
& \rotatebox{90}{Cifar100} & \rotatebox{90}{Caltech101} & \rotatebox{90}{DTD} & \rotatebox{90}{Flowers102} & \rotatebox{90}{Pets} & \rotatebox{90}{SVHN} & \rotatebox{90}{Sun397} & \rotatebox{90}{Camelyon} & \rotatebox{90}{EuroSAT} & \rotatebox{90}{Resisc45} & \rotatebox{90}{Retinopathy} & \rotatebox{90}{Clevr-Count} & \rotatebox{90}{Clevr-Dist} & \rotatebox{90}{DMLab} & \rotatebox{90}{KITTI-Dist} & \rotatebox{90}{dSpr-Loc} & \rotatebox{90}{dSpr-Ori} & \rotatebox{90}{sNORB-Azim\ } & \rotatebox{90}{NsORB-Ele} & \rotatebox{90}{Mean Acc.} \\
\middleline
Full & 68.9 & 87.7 & 64.3 & 97.3 & 86.9 & 87.4 & 38.8   & 79.7 & 95.7 & 84.2 & 73.9     & 56.3 & 58.6 & 41.7 & 65.5 & 57.5 & 46.7 & 25.7 & 29.1 & 68.9 \\
Linear & 64.4 & 85.0 & 63.2 & 97.0 & 86.3 & 36.6 & 51.0 & 78.5 & 87.5 & 68.5 & 74.0 & 34.3 & 30.6 & 33.2 & 55.4 & 12.5 & 20.0 & 9.6 & 19.2 & 57.6 \\
\hline
VPT-Deep    & 78.8 & 90.8 & 65.8 & 98.0 & 88.3 & 78.1 & 49.6 & 81.8 & 96.1 & 83.4 & 68.4 & 68.5 & 60.0 & 46.5 & 72.8 & 73.6 & 47.9 & 32.9 & 37.8 & 72.0 \\
Adapter     & 69.2 & 90.1 & 68.0 & 98.8 & 89.9 & 82.8 & 54.3 & 84.0 & 94.9 & 81.9 & 75.5 & 80.9 & 65.3 & 48.6 & 78.3 & 74.8 & 48.5 & 29.9 & 41.6 & 73.9 \\
AdaptFormer & 70.8 & 91.2 & 70.5 & 99.1 & 90.9 & 86.6 & 54.8 & 83.0 & 95.8 & 84.4 & 76.3 & 81.9 & 64.3 & 49.3 & 80.3 & 76.3 & 45.7 & 31.7 & 41.1 & 74.7 \\
LoRA        & 67.1 & 91.4 & 69.4 & 98.8 & 90.4 & 85.3 & 54.0 & 84.9 & 95.3 & 84.4 & 73.6 & 82.9 & 69.2 & 49.8 & 78.5 & 75.7 & 47.1 & 31.0 & 44.0 & 74.5 \\
NOAH        & 69.6 & 92.7 & 70.2 & 99.1 & 90.4 & 86.1 & 53.7 & 84.4 & 95.4 & 83.9 & 75.8 & 82.8 & 68.9 & 49.9 & 81.7 & 81.8 & 48.3 & 32.8 & 44.2 & 75.5 \\
RepAdapter  & 69.0 & 92.6 & \textbf{75.1} & 99.4 & 91.8 & 90.2 & 52.9 & 87.4 & 95.9 & 87.4 & 75.5 & 75.9 & 62.3 & 53.3 & 80.6 & 77.3 & 54.9 & 29.5 & 37.9 & 76.1 \\
RLRR        & 75.6 & 92.4 & 72.9 & 99.3 & 91.5 & 89.8 & 57.0 & 86.8 & 95.2 & 85.3 & 75.9 & 79.7 & 64.2 & 53.9 & 82.1 & 83.9 & 53.7 & 33.4 & 43.6 & 76.7 \\
GLoRA & 76.4 & 92.9 & 74.6 & \textbf{99.6} & \textbf{92.5} & 91.5 & 57.8 & 87.3 & \textbf{96.8} & 88.0 & 76.0 & 83.1 & 67.3 & 54.5 & \textbf{86.2} & 83.8 & 52.9 & 37.0 & 41.4 & 78.0 \\
\hline
Baseline & 74.9 & 93.3 & 72.0 & 99.4 & 91.0 & 91.5 & 54.8 & 83.2 &  95.7 & 86.9 & 74.2 & 83.0 & 70.5 & 51.9 & 81.4 & 77.9 & 51.7 & 33.6 & 44.4 & 76.4\\
\rowcolor{tblcolor}\ +PACE & \textbf{79.0} & \textbf{94.2} & 73.6 & 99.4 & 92.4 & \textbf{93.7} & \textbf{58.0} & \textbf{87.4} & 96.4 & \textbf{89.3} & \textbf{77.1} & \textbf{84.9} & \textbf{70.9} & \textbf{54.9} & 84.3 & \textbf{84.7} & \textbf{57.3} & \textbf{39.3} & \textbf{44.8} & \textbf{79.0} \\
\bottomline
\end{tabular}
\end{center}
\vspace{-0.5cm}
\end{table}

\section{Experiments}
\label{sec:experiments}
We combine \loramul and \vptadd\  to form a strong baseline \baseline, outperforming other combinations in most cases. We evaluate our method across four visual classification adaptation tasks: VTAB-1K \cite{vtab}, few-shot learning \cite{jie2023fact}, FGVC \cite{vpt} and domain adaptation \cite{noah}. We demonstrate \our improves LoRA on GLUE \cite{wang2018glue} for text classification and GSM-8K \cite{cobbe2021training} for text generation.

\noindent\textbf{Datasets and evluations.} \textbf{VTAB-1K} comprises 19 datasets organized into (i) Natural images, (ii) Specialized datasets (remote sensing, medical) and (iii) Structured datasets (scene structure) domains. Each dataset has 1K training examples. Following \cite{vtab, vpt}, we use the provided 800-200 train split for hyperparameter selection, evaluate using the full training set and report average accuracy across three trails. \textbf{Few-shot learning} involves 5 fine-grained datasets: FGVC-Aircraft \cite{fgvc-aircraft}, Food101 \cite{food101}, OxfordFlowers102 \cite{oxford-flowers}, OxfordPets \cite{oxford-pets} and StanfordCars \cite{stanford-cars}. Following \cite{jie2023fact}, we evaluate 1, 2, 4, 8 and 16 shots, train on the provided training set, tune hyperparameters using validation and report average test accuracy over three random seeds. \textbf{FGVC} includes 5 fine-grained datasets: CUB-200-2011 \cite{WahCUB_200_2011}, NABirds \cite{van2015building}, OxfordFlowers \cite{oxford-flowers}, StanfordDogs \cite{stanford-dogs} and StanfordCars \cite{stanford-cars}. We follow \cite{vpt} to use validation set for hyperparameter and report test results. For \textbf{domain adaptation}, following \cite{noah, glora}, we train on ImageNet \cite{imagenet} with a 16-shot setting, use the validation split by \cite{noah} for hyperparameter selection and report the results on the official validation set and 4 out-of-domain datasets: ImageNet-Sketch \cite{imagenetsketch}, ImageNet-V2 \cite{imagenetv2}, ImageNet-A \cite{imagenetadv} and ImageNet-R \cite{imagenetr}. We evaluate on GLUE \cite{wang2018glue} for \textbf{text classification} and GSM-8K \cite{cobbe2021training} for \textbf{mathematical reasoning}.

\noindent\textbf{Pre-trained backbones}. We experiment with two vision transformers, Vision Transforms (ViT-B/16) \cite{vit} and Swin Transformer (Swin-B) \cite{swin}. These two are pre-trained on ImageNet-21K \cite{imagenet}. We test a ViT-B-Laion-IN12K model, pre-trained on Laion-2B \cite{laion} and fine-tuned on ImageNet-12K \cite{imagenet}. We use RoBERTa$_\text{base}$ \cite{liu2019roberta} and Phi-3-mini-4k-instruct \cite{abdin2024phi} for text classification and generation.

\noindent\textbf{Implementation details}. We follow \cite{vpt} for image processing:  $224\times224$ resizing for VTAB-1K; random flips and crops to $224\times224$ for FGVC and few-shot learning; stronger augmentation for domain adaptation task, following \cite{vit, noah, Lian_2022_SSF}. We use the Adam optimizer \cite{adam} with cosine learning rate decay and linear warm-up (first 10 epochs). Models are fine-tuned for 300 epochs on VTAB-1K and 100 epochs on other vision adaptation tasks, with batch size 64. For text classification we follow \cite{hu2022lora}. See \S\ref{sup:sec:gsm8k} for mathematical reasoning details. All experiments used an NVIDIA H100 GPU.

\noindent\textbf{Baseline.} For each dataset, we identified the better method (\baseline or \loraadd) and tuned the rank, learning rate, and weight decay to form a strong baseline. The detailed baseline settings for each task and the number of trainable parameters are provided in \S \ref{supp:baseline}, where \baseline generally outperformed other variants. Building on the strong \baseline, we use the grid search for our $\lambda$ and $\sigma$, following strategies from previous studies \cite{vpt, Lian_2022_SSF, hu2022lora}. Beyond \baseline, PACE also enhances PEFT methods such as  AdaptFormer, GLoRA, COFT, and BOFT (\S\ref{sup:subsec:other-peft}).

\begin{table}[t]
\vspace{-0.2cm}
\begin{center}
\fontsize{8}{9}\selectfont
\setlength{\tabcolsep}{0.138cm}
\caption{Classification accuracy on Few-shot learning with ViT-B/16 pre-trained on ImageNet-21K.}
\label{tab:fewshot-comparison}
\begin{tabular}{!{\vrule width \boldlinewidth}l|ccccc|ccccc|ccccc!{\vrule width \boldlinewidth}}
\topline
\multirow{2}{*}{\diagbox{Method}{Shot}}
 & \multicolumn{5}{c|}{FGVCAircraft} & \multicolumn{5}{c|}{Food101} & \multicolumn{5}{c!{\vrule width \boldlinewidth}}{Flowers102} \\
\cline{2-16}
& 1 & 2 & 4 & 8 & 16 & 1 & 2 & 4 & 8 & 16 & 1 & 2 & 4 & 8 & 16 \\
\middleline
\loraadd & 10.4 & 15.2 & 27.2 & 41.7 & 59.2 & 33.9 & 51.9 & 59.3 & 66.0 & 71.3 & 93.3 & 96.4 & 98.0 & 98.6 & 98.7  \\
\rowcolor{tblcolor}\ +PACE & 10.7 & 16.3 & 28.2 & 42.1 & 61.0 & 40.6 & 55.9 & 63.8 & 70.3 & 75.2 & 95.0 & 98.0 & 98.9 & 99.5 & 99.6 \\
\hline
\vptadd & 11.2 & 15.1 & 23.7 & 36.3 & 51.5 & 34.3 & 56.6 & 64.8 & 71.7 & 75.4 & 94.3 & 97.6 & 98.2 & 99.3 & 99.6 \\
\rowcolor{tblcolor}\ +PACE & 11.6 & 16.2 & 24.0 & 37.0 & 52.4 & \textbf{39.9} & 57.2 & 66.7 & 72.4 & 76.1 & \textbf{95.3} & 97.8 & 98.6 & 99.4 & 99.6\\
\hline
\baseline & 10.5 & 15.6 & 28.4 & 44.8 & 61.8 & 35.4 & 54.3 & 64.8 & 72.1 & 76.4 & 90.4 & 97.3 & 98.4 & 99.4 & 99.5 \\
\rowcolor{tblcolor}\ +PACE & \textbf{12.3} & \textbf{16.8} & \textbf{29.9} & \textbf{45.7} & \textbf{62.5} & 39.3 & \textbf{57.2} & \textbf{66.7} & \textbf{73.4} & \textbf{77.8} & 93.4 & \textbf{98.1} & \textbf{99.1} & \textbf{99.5} & \textbf{99.7} \\
\middleline
& \multicolumn{5}{c|}{OxfordPets} & \multicolumn{5}{c|}{StanfordCars} & \multicolumn{5}{c!{\vrule width \boldlinewidth}}{Average} \\
\middleline
\loraadd & 73.2 & 83.1 & 87.5 & 89.2 & 91.1 & 8.7 & 15.3 & 30.2 & 55.3 & 74.5 & 43.9 & 52.3 & 60.4 & 70.1 & 78.9 \\
\rowcolor{tblcolor}\ +PACE & 75.3 & 85.0 & \textbf{90.7} & 90.8 & 92.4 & 9.4 & 16.0 & 30.9 & 56.1 & 75.9 & 46.2 & 54.2 & 62.5 & 71.7 & 80.8 \\
\hline
\vptadd  & 75.9 & 85.6 & 90.3 & 90.6 & 92.3 & 9.3 & 15.0 & 27.8 & 46.6 & 65.1 & 45.0 & 53.9 & 60.9 & 68.9 & 76.7\\
\rowcolor{tblcolor}\ +PACE & \textbf{78.2} & 87.4 & 90.3 & 91.1 & 92.3 & \textbf{9.9} & 15.4 & 27.9 & 47.0 & 65.9 & \textbf{46.9} & 54.8 & 61.5 & 69.3 & 77.2 \\
\hline
\baseline & 69.9 & 84.1 & 89.1 & 91.3 & 91.9 & 9.0 & 16.3 & 32.7 & 59.0 & 76.4 & 43.0 & 53.5 & 62.6 & 73.2 & 81.2 \\
\rowcolor{tblcolor}\ +PACE & 76.5 & \textbf{88.0} & 90.3 & \textbf{91.4} & \textbf{92.4} & 9.7 & \textbf{16.4} & \textbf{33.7} & \textbf{59.8} & \textbf{77.3} & 46.2 & \textbf{55.3} & \textbf{63.9} & \textbf{73.9} & \textbf{81.9}\\
\bottomline
\end{tabular}
\end{center}
\vspace{-0.8cm}
\end{table}

\begin{minipage}[h]{\linewidth}
    \begin{minipage}{0.48\textwidth}
\fontsize{8}{9}\selectfont
\setlength{\tabcolsep}{0.05cm}
    \centering
    \captionof{table}{Results on FGVC with ViT-B/16.\\ * denotes using augmented ViT by AugReg \cite{steiner2022how}.}
    \vspace{-0.2cm}
    \label{tab:fgvc-comparison}
    \begin{tabular}{!{\vrule width \boldlinewidth}l|cccccc!{\vrule width \boldlinewidth}}
    \topline
    \multirow{2}{*}{Method} & CUB & NA- & Oxford & Stan. & Stan. & Mean\\
    & -2011 & Birds & Flowers & Dogs & Cars & Acc.\\
    \middleline
    Full & 87.3 & 82.7 & 98.8 & 89.4 & 84.5 & 85.9 \\
    Linear & 85.3 & 75.9 & 97.9 & 86.2 & 51.3 & 79.3 \\
    VPT & 88.5 & 84.2 & 99.0 & 90.2 & 83.6 & 89.1 \\
    LoRA & 88.3 & 85.6 & 99.2 & 91.0 & 83.2 & 89.5 \\
    SSF* & 89.5 & 85.7 & 99.6 & 89.6 & 89.2 & 90.7 \\
    ARC*  &  89.3 & 85.7 & \textbf{99.7} & 89.1 & \textbf{89.5} & 90.7 \\
    RLRR* & 89.8 & 85.3 & 99.6 & 90.0 & 90.4 & 91.0 \\
    \hline
    \baseline & 88.9 & 87.1 & 99.4 & 91.2 & 87.5 & 90.8 \\
    \rowcolor{tblcolor}\ +PACE & \textbf{89.8} & \textbf{87.3} & 99.5  & \textbf{92.2} & 88.8 & \textbf{91.5} \\
    \bottomline
    \end{tabular}
    \end{minipage}
    \hspace{0.012\linewidth}
    \begin{minipage}{0.49\linewidth}
        \fontsize{8}{9}\selectfont
\setlength{\tabcolsep}{0.05cm}
    \centering
    \captionof{table}{Results on domain adaptation with ViT-B/16 pre-trained on ImageNet-21K.}
     \vspace{-0.2cm}
    \label{tab:domain-comparison}
    \begin{tabular}{!{\vrule width \boldlinewidth}l|c|cccc|c!{\vrule width \boldlinewidth}}
    \topline
    \multirow{2}{*}{Method} & Source & \multicolumn{4}{c|}{Target} & Mean\\
    \cline{2-6}
    & ImageNet & -Sketch & -V2 & -A & -R & Acc.\\
    \middleline
    Full & 63.9 & 18.5 & 52.5 & 3.2 & 21.2 & 31.8 \\
    Linear & 67.9 & 14.4 & 60.8 & 9.4 & 25.6 & 35.6 \\
    Adapter & 70.5 & 16.4 & 59.1 & 5.5 & 22.1 & 34.7\\
    VPT & 70.5 & 18.3 & 58.0 & 4.6 & 23.2 & 34.7 \\
    LoRA & 70.8 & 20.0 & 59.3 & 6.9 & 23.3 & 36.0 \\
    NOAH & 71.5 & 24.8 & 66.1 & 11.9 & 28.5  & 40.5 \\
    GLoRA & 78.3 & 30.6 & 67.5 & 13.3 & 31.0 & 44.1 \\
    \hline
    \baseline & 78.3 & 30.6 & 68.5 & 14.1 & 32.5 & 44.8 \\
    \rowcolor{tblcolor}\ +PACE & \textbf{79.0} & \textbf{31.8} & \textbf{69.4} & \textbf{16.3} & \textbf{35.2} & \textbf{46.3}\\
    \bottomline
    \end{tabular}
    \end{minipage}
    \vspace{0.2cm}
\end{minipage}

\begin{minipage}{\linewidth}
\begin{minipage}{0.67\linewidth}
\fontsize{8}{9}\selectfont
\setlength{\tabcolsep}{0.24cm}
    \centering
    \captionof{table}{Results for GLUE w/ RoBERTa$_\text{base}$. Matthew's correlation for COLA, Pearson correlation for STSB, and accuracy for others.}
    \label{tab:gleu}
    \vspace{-0.2cm}
    \begin{tabular}{!{\vrule width \boldlinewidth}l|cccccc|c!{\vrule width \boldlinewidth}}
    \topline
    Method & COLA & STSB & MRPC & RTE  & QNLI & SST2 & Avg.\\
    \hline
    Full   & 63.6 & 91.2 & 90.2 & 78.7 & 92.8 & 94.8 & 85.2\\
    BitFit & 62.0 &	90.8 & \textbf{92.7} & 81.5 & 91.8 & 93.7 & 85.4\\
    Adapt  & 62.6 & 90.3 & 88.4	& 75.9 & 93.0 & 94.7 & 84.2\\
    VeRA   & 65.6 &	90.7 & 89.5	& 78.7 & 91.8 &	94.6 & 85.2\\
    \hline
    LoRA   & 63.4 & 91.5 & 89.7	& 86.6 & 93.3 &	95.1 & 86.6\\
    \rowcolor{tblcolor}\ +\our  & \textbf{66.2} &	\textbf{92.0} & 91.4 & \textbf{86.9} & \textbf{93.6} & \textbf{95.6} &  \textbf{87.6}\\
    \bottomline
    \end{tabular}
\end{minipage}\hfill
\begin{minipage}{0.29\linewidth}
\fontsize{8}{9}\selectfont
\setlength{\tabcolsep}{0.4cm}
\renewcommand{\arraystretch}{1.37}
\centering
\captionof{table}{Results for GSM-8K using Phi-3-mini-4k-instruct.}
\label{tab:gsm8k}
\vspace{-0.2cm}
\begin{tabular}{!{\vrule width \boldlinewidth}l|c!{\vrule width \boldlinewidth}}
\topline
Method & Accuracy \\
\middleline
Pre-trained	& 62.01 \\
Full	& 73.16 \\
\hline
LoRA	& 75.66 \\
\rowcolor{tblcolor}\ +\our	& \textbf{78.77} \\
\bottomline
\end{tabular}
\end{minipage}
\end{minipage}

\begin{table}[h]
\vspace{-0.3cm}
\fontsize{8}{9}\selectfont
\setlength{\tabcolsep}{0.052cm}
    \centering
    \caption{Classification results on domain adaptation and CIFAR-100 in VTAB-1K based different pre-trained models. Src. is short for `source' in Table \ref{tab:domain-comparison}.}
    \label{tab:other-backbones}
    \begin{tabular}{!{\vrule width \boldlinewidth}l|c|ccccc|c|ccccc|c|ccccc!{\vrule width \boldlinewidth}}
    \topline
    \multirow{3}{*}{Method} & \multicolumn{6}{c|}{ViT-B (ImageNet-21K)} & \multicolumn{6}{c|}{ViT-B (Laion2B-ImageNet-12K)} & \multicolumn{6}{c!{\vrule width \boldlinewidth}}{Swin-B (ImageNet-21K)} \\
    \cline{2-19}
    & CIFAR & \multicolumn{5}{c|}{ImageNet-1K} & CIFAR & \multicolumn{5}{c|}{ImageNet-1K} & CIFAR & \multicolumn{5}{c!{\vrule width \boldlinewidth}}{ImageNet-1K}\\
    & -100 & Src. & -S & -V & -A & -R & -100 & Src. & -S & -V & -A & -R & -100 & Src. & -S & -V & -A & -R \\ 
    \middleline
    Full & 51.6 &  63.9  & 18.5 &  52.5 &  3.2 &  21.2 & 51.2 &  66.0 & 29.0 &  56.1 & 8.1 & 27.9 &  65.6 & 71.7 & 27.0 & 61.1 & 10.8 & 24.4 \\
    Linear & 63.4 & 67.9 & 14.4 & 60.8 & 9.4 & 25.6 & 61.9 & 79.2 & 43.2 & 69.5 & 23.4& 40.9 & 65.0 & 78.8 & 36.7 & 68.8 & 23.2 & 35.9  \\
    \loraadd   & 71.2 & 73.8 & 27.1 & 64.8 & 13.6 & 25.0 & 71.3 & 77.5 & 39.8 & 67.8 & 20.4 & 35.6 & 74.3& 76.3& 30.7 & 65.7 & 16.8 & 28.9\\
    \vptadd    & 73.6 & 74.3 & 27.1 & 65.9 & 11.5 & 26.7 & 71.8 & 78.4 & 40.4 & 68.7 & 22.4 & 38.4 & 72.7 & 76.2 & 30.6 & 66.2 & 17.6 & 29.1 \\
    \loramul & 73.4 & 78.1 & 31.2 & 68.3 & 13.4 & 32.7 & 73.2 & 78.6 & 41.9 & 68.8&22.6&37.8& 73.9 & 76.1 & 30.8 & 65.7 & 18.1 & 28.9 \\
    LoRA$_\text{add}$+VPT$_\text{add}$ & 70.3 & 76.8 & 28.7 & 66.6 & 13.7& 29.9 & 71.8 & 78.0 & 41.4 & 68.3 & 20.6 & 36.9 & 74.5 & 76.3 & 30.7 & 65.7 & 16.8 & 28.9 \\
    \hline
    \baseline & 74.9 & 78.3 & 30.6 & 68.5 & 14.1 & 32.5  & 73.8 & 78.3 & 41.5 & 68.6 & 21.6 & 38.2 & 74.6  & 76.6 & 31.2 & 66.5 & 18.5 & 29.4 \\
    \rowcolor{tblcolor}\ +PACE & \textbf{79.0} & \textbf{79.0} & \textbf{31.8} & \textbf{69.4} & \textbf{16.3} & \textbf{35.2} & \textbf{78.0} & \textbf{80.1} & \textbf{45.8} & \textbf{71.2} & \textbf{24.6} & \textbf{43.6} & \textbf{78.9} & \textbf{79.6} & \textbf{39.2} & \textbf{70.1} & \textbf{25.2} & \textbf{38.0} \\
    \bottomline
    \end{tabular}
    \vspace{-0.2cm}
\end{table}

\subsection{Comparison with the State of the Arts}
\noindent\textbf{Results on VTAB-1K.} Table \ref{tab:vtab-comparison} presents the results comparing \our with recent state-of-the-art PEFT methods. \our improves the strong baseline by 2.6\% accuracy, surpassing the previous SOTA GLoRA \cite{glora} by 1\%, which uses two stages for parameter search. In \S\ref{sup:subsec:100epochs}, we show that reducing training epochs to 50 or 100 has minimal impact on PACE performance.

\noindent\textbf{Results on Few-shot Learning}. Table \ref{tab:fewshot-comparison} compares performance w/ and w/o our \our. \our improves \loraadd, \vptadd, \baseline, with \baseline+\our performing best in most cases. \our yields notable improvement, especially when the number of shot is small.

\noindent\textbf{Results on FGVC.} 
Table \ref{tab:fgvc-comparison} shows that \our improves the strong \baseline by 0.7\%, outperforming SSF \cite{Lian_2022_SSF}, ARC \cite{dong2024efficient} and RLRR \cite{dong2024low} that use strongly pre-trained ViT with augmentations. In \S\ref{sup:subsec:fgvc}, PACE achieves larger improvements on smaller datasets.

\noindent\textbf{Results on domain adaptation.} Table \ref{tab:domain-comparison} compares \our with others. \baseline outperforms GLoRA \cite{glora} which relies on parameter search. Meanwhile, \our improves \baseline by 1.5\%, outperforming other PEFT methods, demonstrating superior performance on domain adaptation.

\noindent\textbf{Results on text classification and mathematical reasoning.} Table \ref{tab:gleu} shows that PACE outperforms LoRA by 1\% on GLUE text classification and by 3.11\% on GSM-8K mathematical reasoning.

\noindent\textbf{Generalization on other backbones.} We evaluate \our on CIFAR-100 (VTAB-1K) and domain adaptation using Swin-B \cite{swin} pre-trained on ImageNet-21K and ViT-B (pre-trained on Laion 2B, then fine-tuned on ImageNet-12K). Table \ref{tab:other-backbones} shows \our outperforms \baseline and other PEFT methods across all backbones, demonstrating its strong generalizability. Further experiments in \S\ref{sup:subsec:self-supervised} show PACE works effectively with self-supervised models such as MAE \cite{he2022masked} and DINO \cite{caron2021emerging}.

\subsection{Analyses}
To verify our theories, we conduct experiments on CIFAR-100 (VTAB-1K) using ViT-B/16 and Camelyon (VTAB-1K) on Swin-B. Figures \ref{fig:analysis} \& \ref{fig:analysis-swin} show the gradient norm (summed across all layers) and FP-distance (Eq. \ref{eq:fp_distance}) and the train \& validation accuracy during training for baseline \baseline and \our on validation set. Figures \ref{subfig:compare_gn} \& \ref{subfig:compare_gn_swin} show that \our has a smaller gradient norm than baseline, verifying Theorem \ref{thm:pace} that \our can implicitly lower the weight gradient norm for better generalization. Figures \ref{subfig:compare_dist} \& \ref{subfig:compare_dist_swin} demonstrate that \our maintains a lower FP-distance than the baseline, verifying Theorem \ref{thm:pace_compare} that PACE can implicitly align the fine-tuned model with pre-trained model, retaining knowledge  from large-scale pre-training. Owing to the advantages of the gradient regularization and model alignment, \our shortens the performance gap between seen and unseen data, yielding higher accuracy on the unseen validation set, as shown in Figures \ref{subfig:compare_acc} \& \ref{subfig:compare_acc_swin}.

\begin{figure}[h]
    \centering
    \includegraphics[width=\linewidth]{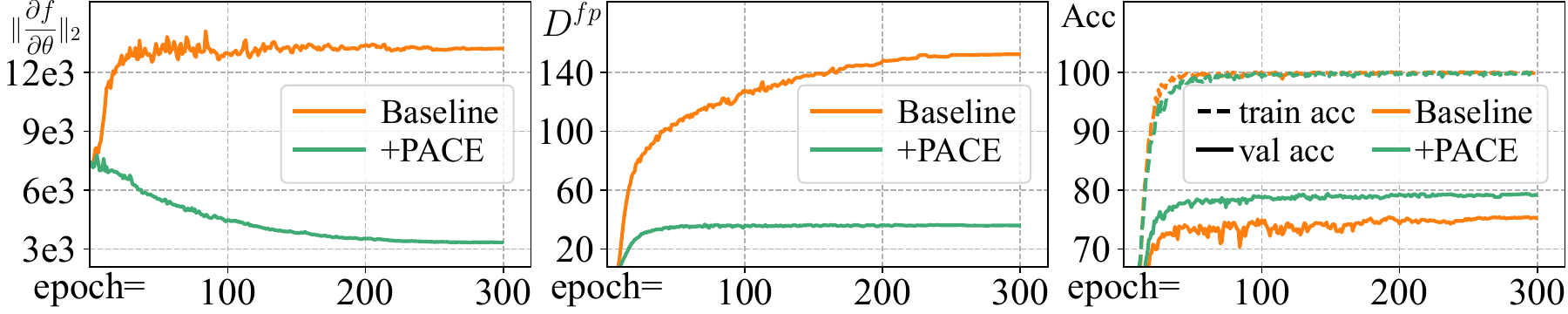}
    \vspace{-0.2cm}
    \begin{subfigure}{0.325\linewidth}
        \caption{Gradient Norm.}\label{subfig:compare_gn}
    \end{subfigure}
    \begin{subfigure}{0.325\linewidth}
        \caption{FP-Distance}\label{subfig:compare_dist}
    \end{subfigure}
    \begin{subfigure}{0.325\linewidth}
        \caption{Train and validation accuracy.}\label{subfig:compare_acc}
    \end{subfigure}
    \caption{Analysis for \our. (a) gradient norm, (b) FP-Distance and (c) train \& val. accuracy are evaluated on validation set of CIFAR-100 (VTAB-1K) with baseline \baseline on ViT-B/16.}
    \label{fig:analysis}
    \vspace{-0.2cm}
\end{figure}

\begin{figure}[h]
    \centering
    \includegraphics[width=\linewidth]{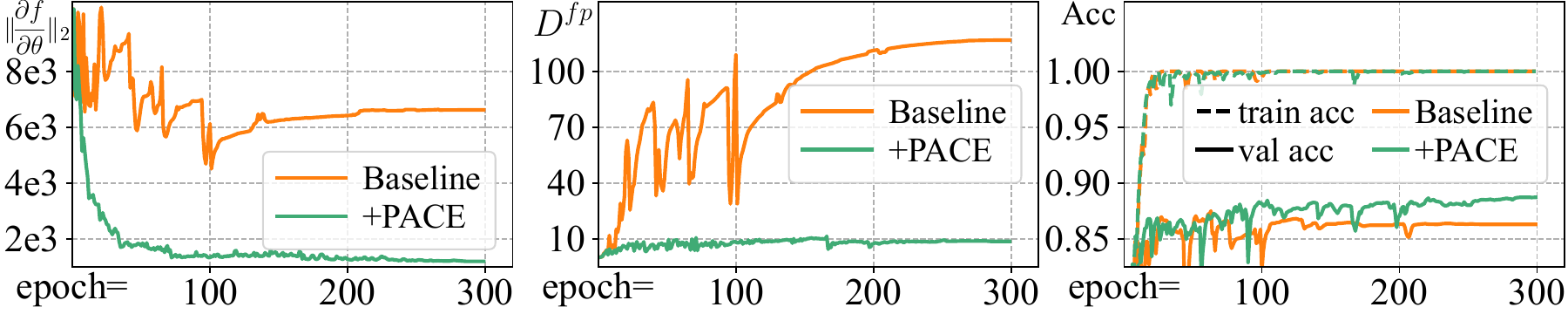}
    \vspace{-0.2cm}
    \begin{subfigure}{0.325\linewidth}
        \caption{Gradient Norm.}\label{subfig:compare_gn_swin}
    \end{subfigure}
    \begin{subfigure}{0.325\linewidth}
        \caption{FP-distance}\label{subfig:compare_dist_swin}
    \end{subfigure}
    \begin{subfigure}{0.325\linewidth}
        \caption{Train and validation accuracy.}\label{subfig:compare_acc_swin}
    \end{subfigure}
    \caption{Analysis for \our. (a) gradient norm, (b) FP-Distance and (c) train \& val. accuracy are evaluated on the validation set of Camelyon (VTAB-1K) with baseline \baseline on Swin-B.}
    \label{fig:analysis-swin}
    \vspace{-0.cm}
\end{figure}

To clarify why naive alignment is problematic, we vary the regularization strength $\lambda$ over a wide range (1e-3 to 5e4) for both Fine-tuned Pre-trained model Alignment (FPA) by minimizing $\Dfp$ in Eq. \ref{eq:fp_distance} and \our. Figure \ref{fig:analysis_gn} shows the averaged gradient norm over training (see also Figures \ref{sup:fig:gradient_norm_training} \& \ref{sup:fig:gn_lbds_swin} for more visualizations). \our robustly lowers gradient norms with larger $\lambda$, while FPA exhibits unpredictable behavior, even causing gradient explosion. This verifies Prop. \ref{prop:fp} that minimizing $\Dfp$ is problematic for gradient regularization, complicating gradient management.

\begin{figure}[h]
    \begin{minipage}{0.64\linewidth}
    \includegraphics[width=\linewidth]{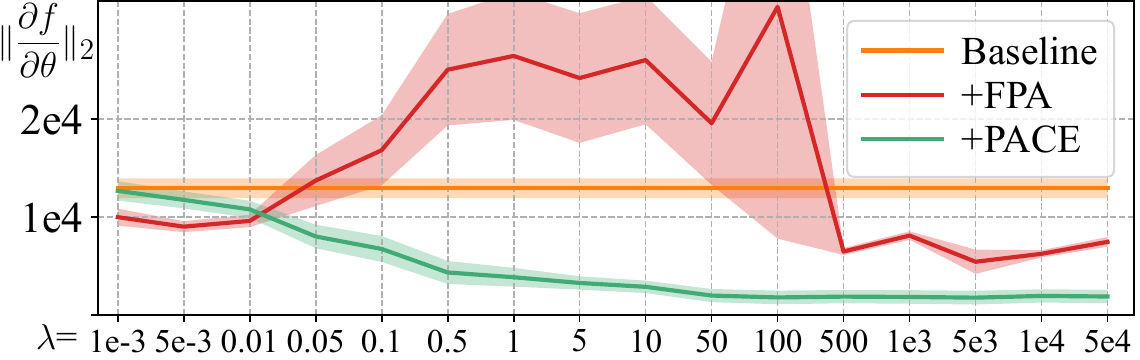}\\
    \vspace{-0.5cm}
    \captionof{figure}{Gradient norms of models across wide range of regularization strengths $\lambda$ on CIFAR-100 (VTAB-1K) w/ ViT-B/16. Line and shadow represent mean and std across training epochs.}
    \label{fig:analysis_gn}
    \end{minipage}
    \hfill
    \begin{minipage}{0.33\linewidth}
    \includegraphics[width=\linewidth]{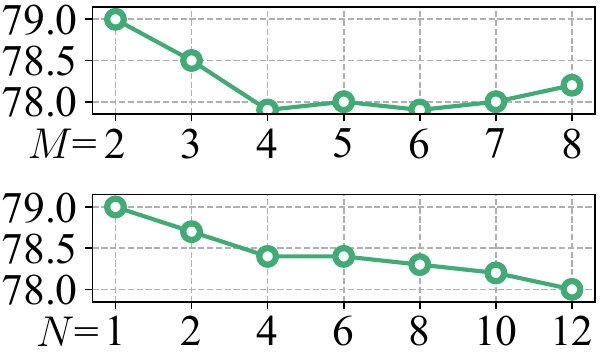}\\
    \vspace{-0.4cm}
    \captionof{figure}{Ablation results for applying \our among $M$ nets and lazily at every $N$ steps.}
    \label{fig:ablation-network-step}
    \end{minipage}
    \vspace{-0.3cm}
\end{figure}

\subsection{Ablation studies}
We ablate \our based on the baseline \baseline on CIFAR-100 (VTAB-1K) and ImageNet-1K in domain adaption as shown in Table \ref{tab:ablation}. The ablations include Noise (baseline w/ noise perturbing adapter), PACE$_\text{add}$ (replacing the multiplicative noise with the additive noise), PACE$_h$ (perturbing $h(\cdot)$ instead of $\Delta h(\cdot)$ in Eq. \ref{eq:pace_noise}), PACE$_\text{drop}$ (replacing the Gaussian noise with the dropout noise), PACE$_{\sigma=}$ (all transformer blocks share the same $\sigma$), PACE$_{\sigma\uparrow}$ ($\sigma$ increases linearly with depth), FPA (fine-tuned and pre-trined alignment by minimizing Eq. \ref{eq:fp_distance}), SAM (sharpness-aware minimization \cite{foret2021sharpnessaware}), GP (gradient penalization), $\ell_1$ (sparsity regularization), and transfer learning methods L2SP \cite{xuhong2018explicit}, DELTA \cite{li2019delta} and FTP \cite{tian2024fast}. We grid-search hyperparameters and report the best results.

Table \ref{tab:ablation} presents the results for all variants. \our improves over Noise, which itself is better than baseline, justifying our adapter perturbation and consistency regularization. PACE$_\text{add}$ performs worse than \our, showing the superiority of the multiplicative noise. Although PACE$_h$ can implicitly regularize gradients, it performs worse than \our, verifying the advantages of perturbing adapter to implicitly align models. PACE$_\text{drop}$ is worse than \our, indicating the dropout noise is suboptimal. PACE$_{\sigma=}$ and PACE$_{\sigma\uparrow}$ perform worse, justifying our design of linearly decreasing $\sigma$.  FPA, SAM and GP, which either only align models or only regularize gradients, are outperformed by \our. Despite combining FPA+GP, it still performs worse than  ours, suggesting ineffective combination. $\ell_1$, L2SP, DELTA, and FTP obtain worse results than \our, showing their limitations in improving generalization. \our regularizes gradients for better generalization and aligns models to retain knowledge, surpassing all other variants.

\begin{figure}[h]
    \begin{minipage}{0.495\linewidth}
        \fontsize{8}{9}\selectfont
    \renewcommand{\arraystretch}{0.95}
    \setlength{\tabcolsep}{0.05cm}
    \centering
    \begin{tabular}{!{\vrule width \boldlinewidth}l|c|ccccc!{\vrule width \boldlinewidth}}
    \topline
    \multirow{2}{*}{Method} & CIFAR & \multicolumn{5}{c!{\vrule width \boldlinewidth}}{ImageNet-1K} \\
    & -100 & Source & -Sketch & -V2 & -A & -R \\
    \middleline
    \baseline & 74.9 & 78.3 & 30.6 & 68.5 & 14.1 & 32.5\\
    \hline
    \ +Noise & 77.4 & 78.3 & 31.3 & 68.6 & 14.3 & 33.0\\
     \rowcolor{tblcolor}\ +PACE & \textbf{79.0} & \textbf{79.0} & \textbf{31.8} & \textbf{69.4} & 16.3 & \textbf{35.2}\\
    \hline
    \ +PACE$_\text{add}$ & 75.7 & 78.3 & 31.2 & 68.7 & 13.7 & 32.7 \\
    \ +PACE$_h$ 
    & 75.9 & 78.4 & 31.2 & 68.1 & 13.8 & 32.6 \\
    \ +PACE$_\text{drop}$ & 78.3 & 78.9 & 31.2 & 68.9 & 16.0 & 34.6\\
    \ +PACE$_{\sigma=}$ & 77.9 & 78.8 & 31.6 & 68.3 & \textbf{16.6} & 34.7  \\
    \ +PACE$_{\sigma\uparrow}$ & 77.3 & 78.7 & 31.3 & 68.9 & 14.0 & 33.6 \\
    \hline
    \ +FPA & 76.6 & 78.8 & 31.2 & 68.6 & 14.7 & 33.5 \\
    \ +SAM \cite{foret2021sharpnessaware} & 75.4 & 78.4 & 31.4 & 68.5 & 13.8 & 32.9\\
    \ +GP & 75.8 & 78.3 & 31.7 & 68.4 & 14.2 & 32.1 \\
    \ +FPA+GP & 74.9 & 78.1 & 31.5 & 68.1 & 13.5 & 32.6 \\
    \ +$\ell_1$ & 75.2 & 78.2 & 30.6 & 68.6 & 13.7 & 32.8 \\
    \ +L2SP \cite{xuhong2018explicit}	& 75.9	& 78.5	& 30.4	& 68.7	& 14.9	& 33.5 \\
    \ +DELTA \cite{li2019delta} & 76.4	& 78.4	& 30.8	& 68.7	& 14.6	& 33.7 \\
    \ +FTP \cite{tian2024fast}	& 76.2	& 78.6	& 30.8	& 68.6	& 15.8	& 33.6 \\
    \bottomline
    \end{tabular}
    \captionof{table}{Accuracy results on domain adaptation and VTAB-1K based different pre-trained models. }
    \label{tab:ablation}
    \end{minipage}
    \hspace{0.1cm}
    \begin{minipage}{0.495\linewidth}
        \includegraphics[width=\linewidth]{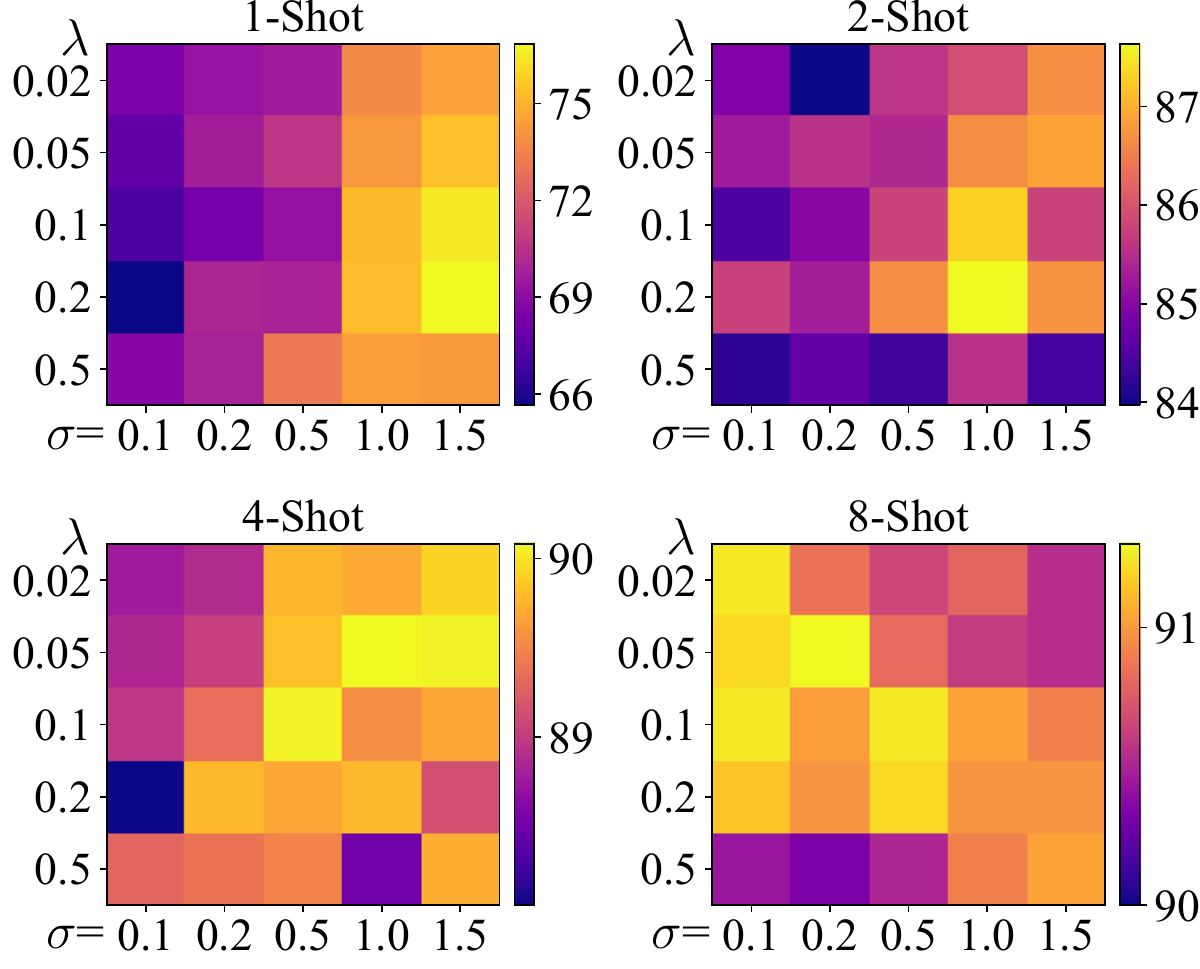}
        \captionof{figure}{Results for varied $\lambda$ and $\sigma$ as well as shot on  OxfordPets in few-shot learning.}
        \label{fig:albation-lambda-sigma}
    \end{minipage}
\end{figure}

We further evaluate applying \our across multiple $M$ networks during training or applying it lazily with half-batch size at every $N$ steps (\pacehalflazy in \S\ref{sup:sec:efficient-pace}). Figure \ref{fig:ablation-network-step} presents the results, showing that applying \our among two networks at every training step performs best. However, lazy regularization applied every few steps can still provide reasonable results while saving computational/memory costs.

We test the sensitivity of hyperparameters $\lambda$ and $\sigma$ introduced in our \our on OxfordPets for few-shot learning across 1, 2, 4, 8 shots. The results presented in Figure \ref{fig:albation-lambda-sigma} demonstrate that with less data, larger $\lambda$ and $\sigma$ are favored, verifying  the effectiveness of \our in improving generalization.

\section{Conclusions}
We have introduced \our, a novel and effective method that combines generalization of PArameter-efficient fine-tuning with Consistency rEgularization. Through rigorous theoretical analyses, we have shown \our reduces weight gradient for improved generalization and it aligns the fine-tuned model with the pre-trained model for retaining pre-training knowledge. Our experimental results support the theoretical analyses, justifying the generalization advantages of \our over other PEFT methods. With its dual advantages, \our consistently outperforms other variants across different backbones, firmly establishing \our as a powerful solution for enhancing generalization for PEFT methods. Limitations and border impacts are discussed in \S \ref{sup:sec:limitations}.

\textbf{Acknowledgments.} We thank Moyang Liu, Melody Ip, Chenyi Du, and Yinuo Xu for their valuable discussions and support. PK is funded by CSIRO’s Science Digital.

{\small
\bibliographystyle{ieee_fullname}
\bibliography{reference.bib}
}

\appendix
\author{%
Yao Ni$^{\dagger}$ \quad Shan Zhang$^{\ddagger,\dagger}$ \quad Piotr Koniusz$\;^{*,\S,\dagger}$\\
$^{\dagger}$The Australian National University \quad $^\S$Data61$\!${\color{red}\heart}CSIRO \\
$^{\ddagger}$Australian Institute for Machine Learning, The University of Adelaide \\
{\tt\small $^{\dagger}$yao.ni@anu.edu.au $^{\ddagger}$shan.zhang@adelaide.edu.au $^\S$piotr.koniusz@data61.csiro.au}
}
\title{PACE: Marrying generalization of PArameter-efficient fine-tuning with Consistency rEgularization (Supplementary Material)}
\makeatletter
\def\ps@myheadings{%
    \let\@oddfoot\@empty\let\@evenfoot\@empty
    \def\@evenhead{\thepage\hfil\slshape\leftmark}%
    \def\@oddhead{{\slshape\rightmark}\hfil\thepage}%
    \let\@mkboth\@gobbletwo
    \let\sectionmark\@gobble
    \let\subsectionmark\@gobble
    }
  \if@titlepage
  \renewcommand\maketitle{\begin{titlepage}%
  \let\footnotesize\small
  \let\footnoterule\relax
  \let \footnote \thanks
  \null\vfil
  \vskip 60\p@
  \begin{center}%
    {\LARGE \@title \par}%
    \vskip 3em%
    {\large
     \lineskip .75em%
      \begin{tabular}[t]{c}%
        \@author
      \end{tabular}\par}%
      \vskip 1.5em%
    {\large \@date \par}
  \end{center}\par
  \@thanks
  \vfil\null
  \end{titlepage}%
  \setcounter{footnote}{0}%
}
\else
\newcommand\maketitle{\par
  \begingroup
    \renewcommand\thefootnote{\@fnsymbol\c@footnote}%
    \def\@makefnmark{\rlap{\@textsuperscript{\normalfont\color{black}\@thefnmark}}}%
    \long\def\@makefntext##1{\parindent 1em\noindent
            \hb@xt@1.8em{%
                \hss\@textsuperscript{\normalfont\@thefnmark}}##1}%
    \if@twocolumn
      \ifnum \col@number=\@ne
        \@maketitle
      \else
        \twocolumn[\@maketitle]%
      \fi
    \else
      \newpage
      \global\@topnum\z@   
      \@maketitle
    \fi
    \thispagestyle{plain}\@thanks
  \endgroup
  \setcounter{footnote}{0}%
}
\makeatother

\maketitle

\section{Broader impacts and limitations}
\label{sup:sec:limitations}

\subsection{Broader impacts} Our work provides a powerful solution for improving generalization in Parameter Efficient Fine-Tuning (PEFT), allowing for effective fine-tuning of pre-trained models while reducing the heavily reliance on pre-training from scratch using large scale data. Our advancements in PEFT, supported by Theorems \ref{thm:grad}, \ref{thm:pace} and \ref{thm:pace_compare}, offer novel insights into gradient regularization and model alignment. These insights extend beyond PEFT and can be applied to other areas such as continual learning and transfer learning, potentially enhancing the performance and efficiency of models in various domains. By leveraging our findings, practitioners can develop more robust and adaptable models that generalize well to new tasks and environments, leading to more intelligent and versatile AI systems. In terms of negative impacts, the robustness of our fine-tuning method could potentially be misused to create more convincing deepfakes, raising concerns about the spread of misinformation, manipulation of public opinion, and malicious activities such as fraud, blackmail, or harassment. However, potential misuse is a downside with any improvements that have universal nature.

\subsection{Limitations}
While our work effectively improves generalization ability, it introduces additional computational costs by requiring input samples to be passed through the network twice for regularization. However, this can be mitigated by using two efficient variants, \pacefast and \pacehalflazy, proposed in \S\ref{sup:sec:efficient-pace}, where we demonstrate the potential for resource-efficient fine-tuning. Additionally, our method introduces extra hyperparameters $\lambda$ and $\sigma$, which require caution during hyperparameter search. Nonetheless, Figure \ref{fig:albation-lambda-sigma} suggests that fewer training data requires larger $\lambda$ and $\sigma$ values, providing insight for hyperparameter tuning.

\section{Proofs}
\label{sup:sec:proof}

\subsection{Proof of Theorem \ref{thm:grad}}
\label{sup:sec:proof_thm_grad}
Settting $\vepsilon=\frac{\rho\nablatheta}{\lVert\nablatheta\lVert_2}$, we perform a second-order Taylor expansion of $\calL_{\calDn}$ around $\vtheta$. By incorporating the higher-order terms from the Taylor expansion into $R\Big(\frac{\lVert\vtheta\rVert_2^2}{\rho^2}, \frac{1}{n}\Big)$, we derive:
\begin{align}
    \calL_{\scrD}(\vtheta)&\leq\calL_{\calDn}\Big(\vtheta+\frac{\rho\nablatheta}{\lVert\nablatheta\lVert_2}\Big) + R\Big(\frac{\lVert\vtheta\rVert_2^2}{\rho^2}, \frac{1}{n}\Big)\nonumber\\
    &\approx\calL_{\calDn}(\vtheta)+\rho\lVert\nablatheta\rVert_2+\frac{\rho^2}{2\lVert\nablatheta\rVert_2^2}\nablatheta^T\Htheta\nablatheta+R\Big(\frac{\lVert\vtheta\rVert_2^2}{\rho^2},\frac{1}{n}\Big)\label{sup:eq:grad_taylor}.
\end{align}
Assuming that the approximation does not alter the inequality relationship, \ie, it preserves the $\leq$ relation on both sides and considering the largest eigenvalue of $\Htheta$ as $\lambdaHmax$, implying $\vv^T\Htheta\vv\leq\lambdaHmax\lVert\vv\rVert_2^2$ for any $\vv$, we further bound Eq. \ref{sup:eq:grad_taylor} as follows and arrive at:
\begin{equation}
    \calL_{\scrD}(\vtheta)\leq\calL_{\calDn}(\vtheta)+\rho\lVert\nablatheta\rVert_2+\frac{\rho^2}{2}\lambdaHmax+R\Big(\frac{\lVert\vtheta\rVert_2^2}{\rho^2}, \frac{1}{n}\Big).\nonumber
\end{equation}
\subsection{Proof of Theorem \ref{thm:pace}}
\label{sup:sec:proof_thm_pace}
The proof is motivated by Ni and Koniusz \cite{ni2024nice}. We include the proof process for completeness. Denote $\vm_1=\vz_1-\vone, \vm_2=\vz_2-\vone$ thus $\vm_1, \vm_2\sim\calN(\vzero, \sigma^2)$
\begin{align}
    \dpace=&\bbE_{\vz_1, \vz_2}[f(\vtheta_0+\vz_1\odot\Delta\vtheta) - f(\vtheta_0+\vz_2\odot\Delta\vtheta)]^2\nonumber\\
        =&\bbE_{\vz_1, \vz_2}[f(\vtheta_0+\Delta\vtheta+(\vz_1-\vone)\odot\Delta\vtheta) - f(\vtheta_0+\Delta\vtheta+(\vz_2-\vone)\odot\Delta\vtheta)]^2\nonumber\\
        =&\bbE_{\vm_1, \vm_2}[f(\vtheta+\vm_1\odot\Delta\vtheta) - f(\vtheta+\vm_2\odot\Delta\vtheta)]^2.\label{sup:eq:pace_taylor}
\end{align}

Defining $\vv:=\vm_1\odot\Delta\vtheta$ and $\vu:=\vm_2\odot\Delta\vtheta$, where $\vv, \vu\sim\calN(\vzero, \sigma^2\text{diag}(\Delta\vtheta\odot\Delta\vtheta))$, we can rewrite Eq. \ref{sup:eq:pace_taylor} as follows:
\begin{align}
    &\bbE_{\vv, \vu}[f(\vtheta+\vv) - f(\vtheta+\vu)]^2\nonumber\\
    \approx&\bbE_{\vv, \vu}\big[f(\vtheta)+\vv^T\vnabla+\frac{1}{2}\vv^T\mH\vv-f(\vtheta)-\vu^T\vnabla-\frac{1}{2}\vu^T\mH\vu\big]^2\nonumber\\
    =&\bbE_{\vv, \vu}\big[\vv^T\vnabla+\frac{1}{2}\vv^T\mH\vv-\vu^T\vnabla-\frac{1}{2}\vu^T\mH\vu\big]^2\nonumber\\
    =&\bbE_{\vv, \vu}\big[(\vv-\vu)^T\vnabla+\frac{1}{2}\vv^T\mH\vv-\frac{1}{2}\vu^T\mH\vu\big]^2 \nonumber\\
    =&\bbE_{\vv, \vu}\big[(\vv-\vu)^T\vnabla\big]^2 \label{sup:eq:pace_taylor_term1}\\
    &+\bbE_{\vv, \vu}\big[\big((\vv-\vu)^T\vnabla\big)\big(\vv^T\mH\vv-\vu^T\mH\vu\big)\big]\label{sup:eq:pace_taylor_term2}\\
    &+\frac{1}{4}\bbE_{\vv}[\vv^T\mH\vv]^2+\frac{1}{4}\bbE_{\vu}[\vu^T\mH\vu]^2\label{sup:eq:pace_taylor_term3}\\
    &-\frac{1}{2}\bbE_{\vv, \vu}\big[(\vv^T\mH\vv)(\vu^T\mH\vu)].\label{sup:eq:pace_taylor_term4}
\end{align}
Next, we derive the four terms, Eq. \ref{sup:eq:pace_taylor_term1}, \ref{sup:eq:pace_taylor_term2}, \ref{sup:eq:pace_taylor_term3}, and \ref{sup:eq:pace_taylor_term4}, respectively as follows:

\textbf{Eq. \ref{sup:eq:pace_taylor_term1}.} Using $\bbE_{z_1,z_2}[(z_1-z_2)^2]=2\sigma^2$ for $z_1, z_2 \sim \calN(0, \sigma^2)$, we can simplify (Eq. \ref{sup:eq:pace_taylor_term1}) as follows, noting that terms related to different dimensions are canceled due to zero-mean independent Gaussian noise:
\begin{align}
    \bbE_{\vv, \vu}\big[(\vv-\vu)^T\vnabla\big]^2=\bbE_{\vv, \vu}\big[\sum_j(v_j-u_j)^2\nabla_j^2\big]=2\sigma^2\sum_j\Delta\theta_j^2\nabla_k^2.\label{sup:eq:pace_taylor_term1_result}
\end{align}

\textbf{Eq. \ref{sup:eq:pace_taylor_term2}.}
Utilizing $E[z^3]=\mu^3 + 3\mu\sigma^2$ for $z \sim \calN(\mu, \sigma^2)$, and noting that $E[z^3] = 0$ for $\mu=0$, Eq. \ref{sup:eq:pace_taylor_term2} is derived as:
\begin{align}
    &\bbE_{\vv, \vu}\big[\big((\vv-\vu)^T\vnabla\big)\big(\vv^T\mH\vv-\vu^T\mH\vu\big)\big]\nonumber\\
    =&\bbE_{\vv}\big[(\vv^T\!\vnabla)(\vv^T\!\mH\!\vv)]\!+\!\bbE_{\vu}\big[(\vu^T\!\vnabla)(\vu^T\!\mH\!\vu)]\!-\!\bbE_{\vv,\vu}\big[(\vv^T\!\vnabla)(\vu^T\!\mH\!\vu)]\!-\!\bbE_{\vv,\vu}\big[(\vu^T\!\vnabla)(\vv^T\!\mH\!\vv)]\nonumber\\
    =&2\bbE_{\vv}\big[(\vv^T\!\vnabla)(\vv^T\!\mH\!\vv)]=0.\label{sup:eq:pace_taylor_term2_result}
\end{align}

\textbf{Eq. \ref{sup:eq:pace_taylor_term3}.} We first decompose Eq. \ref{sup:eq:pace_taylor_term3}, then discuss each case and obtain the final result:
\begin{equation}
    \frac{1}{4}\bbE_{\vv}[\vv^T\mH\vv]^2+\frac{1}{4}\bbE_{\vu}[\vu^T\mH\vu]^2
    =\frac{1}{2}\bbE_{\vv}[\vv^T\mH\vv]^2=\frac{1}{2}\bbE_{\vv}\big[\sum_{j,k,p,q}v_jH_{jk}v_kv_pH_{pq}v_q\big].\label{eq:sup:nice_taylor_term3_middle}
\end{equation}
Given the independence of elements in $\vv$, only terms with an element repeated two or four times contribute non-zero results, leading to four distinct, non-overlapping cases. Using $\bbE[z^2]=\sigma^2+\mu^2$ and $\bbE[z^4]=\mu^4+6\mu^2\sigma^2+3\sigma^4$ for $z \sim \calN(\mu, \sigma^2)$, and simplifying to $\bbE[z^2]=\sigma^2$ and $\bbE[z^4]=3\sigma^4$ when $\mu=0$, we have:

\textit{\textbf{Case 1:}} $j=k\neq p=q$, given the independence of $v_j$ and $v_p$, we have:
\begin{equation}
    \bbE_{\vv}\big[\sum_j\sum_{p\neq j} v_j^2H_{jj}v_p^2H_{pp}\big]=\sum_{j,p\neq j}H_{jj}H_{pp}\bbE[v_j^2]\bbE[v_p^2]=\sigma^4\!\!\sum_{j,k\neq j}\!H_{jj}H_{kk}\Delta\theta_j^2\Delta\theta_k^2.
\end{equation}
\textit{\textbf{Case 2:}} For $j=p\neq k=q$,  the independence of $v_j$ and $v_k$ simplifies our calculation, leading to:
\begin{equation}
    \bbE_{\vv}\big[\sum_j\sum_{k\neq j}v_jH_{jk}v_kv_jH_{jk}v_k\big]=\sum_{j,k\neq j}H_{jk}^2\bbE[v_j^2]\bbE[v_k^2]=\sigma^4\sum_{j,k\neq j}H_{jk}^2\Delta\theta_j^2\Delta\theta_k^2.
\end{equation}
\textit{\textbf{Case 3:}} For $j=q\neq k=p$, utilizing the independence of $v_j$ and $v_k$ as well as the symmetry $H_{jk}=H_{kj}$, we obtain:
\begin{equation}
    \bbE_{\vv}\big[\sum_j\sum_{k\neq j}v_jH_{jk}v_kv_k H_{kj}v_j\big]=\sum_{j,k\neq j}H_{jk}^2\bbE[v_j^2]\bbE[v_k^2]=\sigma^4\sum_{j,k\neq j}H_{jk}^2\Delta\theta_j^2\Delta\theta_k^2.
\end{equation}

\textit{\textbf{Case 4:}} For $j=q=k=p$, using $\bbE[z^4]=3\sigma^4$ where $z \sim \calN(0, \sigma^2)$, we have:
\begin{equation}
    \bbE_{\vv}\Big[\sum_jv_jH_{jj}v_jv_jH_{jj}v_j\Big]=\sum_{j}H_{jj}^2\bbE[v_j^4]=3\sigma^4\sum_{j}H_{jj}^2\Delta\theta_j^4.
\end{equation}

Combining above four cases together, we have the result for Eq. \ref{sup:eq:pace_taylor_term3}:
\begin{equation}
    \frac{\sigma^4}{2}\Big(\sum_j3H_{jj}^2\Delta\theta_j^4+\sum_{j,k\neq j}(H_{jj}H_{kk} + 2H_{jk}^2)\Delta\theta_j^2\Delta\theta_k^2\Big).\label{sup:eq:pace_taylor_term3_result}
\end{equation}

\textbf{Eq. \ref{sup:eq:pace_taylor_term4}:}
\begin{align}
    &-\frac{1}{2}\bbE_{\vv, \vu}\big[(\vv^T\mH\vv)(\vu^T\mH\vu)]\nonumber\\
    =&-\frac{1}{2}\bbE_{\vv}\big[(\vv^T\mH\vv)\big]\bbE_{\vu}\big[(\vu^T\mH\vu)\big]\nonumber\\
    =&-\frac{1}{2}\bbE_{\vv}\big[\sum_jH_{jj}v_j^2\big]\bbE_{\vu}\big[\sum_kH_{kk}v_k^2\big]\nonumber\\
    =&-\frac{1}{2}\Big(\sum_jH_{jj}\bbE[v_j^2]\Big)\Big(\sum_kH_{kk}\bbE[v_k^2]\Big)\nonumber\\
    =&-\frac{\sigma^4}{2}\Big(\sum_jH_{jj}^2\Delta\theta_j^4+\sum_{j,k\neq j}H_{jj}H_{kk}\Delta\theta_j^2\Delta\theta_k^2\Big).\label{sup:eq:pace_taylor_term4_result}
\end{align}
With results of Eq. \ref{sup:eq:pace_taylor_term1_result}, \ref{sup:eq:pace_taylor_term2_result}, \ref{sup:eq:pace_taylor_term3_result}, \ref{sup:eq:pace_taylor_term4_result}, we have the final results:
\begin{align}
    \!\!\!\dpace\approx&2\sigma^2\sum_j\Delta\theta_j^2\nabla_j^2+0\nonumber\\
    &+\!\frac{\sigma^4}{2}\!\Big(\!\sum_j3H_{jj}^2\Delta\theta_j^4\!+\!\!\!\sum_{j,k\neq j}\!\!(\!H_{jj}H_{kk}+2H_{jk}^2)\Delta\theta_j^2\Delta\theta_k^2 -\!\! \sum_j\!\!H_{jj}^2\Delta\theta_j^4 -\!\!\!\sum_{j,k\neq j}\!\!\!H_{jj}H_{kk}\Delta\theta_j^2\Delta\theta_k^2\Big)\nonumber\\
    =&2\sigma^2\sum_j\Delta\theta_j^2\nabla_j^2+\sigma^4\Big(\sum_jH_{jj}^2\Delta\theta_j^4+\sum_{j,k\neq j}H_{jk}^2\Delta\theta_j^2\Delta\theta_k^2\Big)\nonumber\\
    =&2\sigma^2\sum_j\Delta\theta_j^2\nabla_k^2+\sigma^4\sum_{j,k}H_{jk}^2\Delta\theta_j^2\Delta\theta_k^2=2\sigma^2\lVert\Delta\vtheta\odot\vnabla\rVert_2^2+\sigma^4\lVert(\Delta\vtheta\Delta\vtheta^T)\odot\mH\lVert_F^2.
\end{align}

\subsection{Proof of Theorem \ref{thm:pace_compare}}
\label{sup:sec:proof_pace_compare}
The Cauchy-Schwarz inequality states that for $\vu, \vv\in\bbR^d$, we have $(\sum_ju_jv_j)^2\leq(\sum_ju_j^2)(\sum_jv_j^2)$. Let $\vu=\vone$, it follows that
$(\sum_jv_j)^2\leq d\lVert \vv \rVert_2^2$. Using this inequality, we then prove the following:
\begin{equation}
     [\Delta\vtheta^T\vnabla-\frac{1}{2}\Delta\vtheta^T\mH\Delta\vtheta]^2\leq2[\Delta\vtheta^T\vnabla]^2+[\Delta\vtheta^T\mH\Delta\vtheta]^2\nonumber
\end{equation}
\begin{equation}
    [\Delta\vtheta^T\vnabla]^2=\Big(\sum_j\Delta\theta_j\nabla_j\Big)^2\leq d\lVert\Delta\vtheta\odot\vnabla\rVert_2^2.
\end{equation}
\begin{equation}
    [\Delta\vtheta^T\mH\Delta\vtheta]^2=\Big(\sum_{j,k}\Delta\theta_j\Delta\theta_kH_{jk}\Big)^2\leq d^2\big\lVert(\Delta\vtheta\Delta\vtheta^T)\odot\mH\big\rVert_F^2
\end{equation}
Here, the inequality is obtained by treating $\Delta\theta_j\Delta\theta_kH_{jk}$ as an element of a vector with size of $d^2$. This leads to the final results.

\subsection{Rationale for one-dimensional output analysis}
\label{sup:subsec:onedim}
We use the squared $L_2$ distance for multi-dimensional outputs for $\Dfp$ and $\Dpace$, which allows our one-dimensional analysis to naturally generalize to multiple dimensions. For example, for a vector-valued function in the naive alignment, $f(\vtheta)=[f_1(\vtheta),...,f_m(\vtheta)]$, where $m$ is the output dimension, we have:
\begin{equation}
    \lVert f(\vtheta_0)-f(\vtheta_0+\Delta\vtheta)\rVert_2^2=\sum_{i=1}^m[f_i(\vtheta_0)-f_i(\vtheta_0+\Delta\vtheta)]^2.\nonumber
\end{equation}
This equality shows that the squared $L_2$ distance in multiple dimensions is simply the sum of non-negative squared differences in each dimension. Consequently, this additive nature enables our one-dimensional analysis to extend seamlessly to multiple dimensions in practice, aligning with our empirical observations.

\subsection{$\boldsymbol{R}$ increases with $\boldsymbol{\frac{1}{n}}$}
\label{sup:subsec:sample_size}
According to \cite{foret2021sharpnessaware}, the function $R\big(\frac{\lVert\vtheta\rVert_2^2}{\rho^2}, \frac{1}{n}\big)$ in Eq. \ref{eq:sam} is defined as:
\begin{equation}
R\Big(\frac{\lVert\vtheta\rVert_2^2}{\rho^2}, \frac{1}{n}\Big) = \sqrt{\frac{k\log\Big(1+\frac{\lVert\vtheta\rVert_2^2}{\rho^2}\big(1+\sqrt{\frac{\log n}{k}}\big)^2\Big)+4\log \frac{n}{\delta}+8\log(6n+3k)}{n-1}}.\nonumber
\end{equation}
Here $k$ is the number of parameters, $n$ is the number of training samples, $\delta\in(0, 1]$ is the confidence level and $\rho$ is the max norm of the Gaussian perturbation noise.

To ensure $R$ is valid, we require $n>1$. To analyze how $R$ changes with $n$,  we fix $\frac{\lVert\vtheta\rVert_2^2}{\rho^2}$ and break the expression under the square root of $R$ into three terms:
\begin{equation}
    R_1=\frac{k\log\Big(1+\frac{\lVert\vtheta\rVert_2^2}{\rho^2}\big(1+\sqrt{\frac{\log n}{k}}\big)^2\Big)}{n-1}, \quad R_2=\frac{4\log n - 4\log \delta}{n-1}, \quad R_3 = \frac{8\log(6n+3k)}{n-1}\nonumber
\end{equation}
We analyze each term separately to determine whether it decreases with increasing $n$.

\textbf{Analysis for $\boldsymbol{R_1}$:} The derivative for $R_1$ \wrt $n$ is:

\begin{align}
    \!\!R_1'\!&=\frac{\frac{k}{1+\frac{\lVert\vtheta\rVert_2^2}{\rho^2}\!\big(\!1+\sqrt{\!\frac{\log n}{k}}\big)^2}\cdot 2\frac{\lVert\vtheta\rVert_2^2}{\rho^2}(1\!+\!\sqrt{\!\frac{\log n}{k}})\cdot\frac{1}{2\sqrt{\!\frac{\log n}{k}}}\cdot\frac{1}{kn}\cdot(n\!-\!1)\!-\!k\log\!\Big(1\!+\!\frac{\lVert\vtheta\rVert_2^2}{\rho^2}\big(1\!+\!\sqrt{\!\frac{\log n}{k}}\big)^2\Big)}{(n-1)^2}.\nonumber\\
    &=\frac{\frac{\frac{\lVert\vtheta\rVert_2^2}{\rho^2}\big(1+\sqrt{\frac{\log n}{k}}\big)}{1+\frac{\lVert\vtheta\rVert_2^2}{\rho^2}\big(1+\sqrt{\frac{\log n}{k}}\big)^2}\cdot\frac{1}{\sqrt{\frac{\log n}{k}}}\cdot\frac{n-1}{n}-k\log\Big(1+\frac{\lVert\vtheta\rVert_2^2}{\rho^2}\big(1+\sqrt{\frac{\log n}{k}}\big)^2\Big)}{(n-1)^2}\nonumber\\
    &<\frac{\frac{\frac{\lVert\vtheta\rVert_2^2}{\rho^2}\big(1+\sqrt{\frac{\log n}{k}}\big)}{\frac{\lVert\vtheta\rVert_2^2}{\rho^2}\big(1+\sqrt{\frac{\log n}{k}}\big)^2}\cdot\frac{1}{\sqrt{\frac{\log n}{k}}}-k\log\Big(\frac{\lVert\vtheta\rVert_2^2}{\rho^2}\big(1+\sqrt{\frac{\log n}{k}}\big)^2\Big)}{(n-1)^2}\nonumber\\
    &<\frac{\frac{1}{1+\sqrt{\frac{\log n}{k}}}\cdot\frac{1}{\sqrt{\frac{\log n}{k}}}-k\Big(\log\frac{\lVert\vtheta\rVert_2^2}{\rho^2}+\log\big(1+\sqrt{\frac{\log n}{k}}\big)^2\Big)}{(n-1)^2}\nonumber\\
    &<\frac{\frac{1}{\sqrt{\frac{\log n}{k}}}\cdot\frac{1}{\sqrt{\frac{\log n}{k}}}-k\log\frac{\lVert\vtheta\rVert_2^2}{\rho^2}-k\log\big(1+\sqrt{\frac{\log n}{k}}\big)^2}{(n-1)^2}\nonumber\\
    &= \frac{k}{(n-1)^2}\cdot\Big(\frac{1}{\log n}-\log\frac{\lVert\vtheta\rVert_2^2}{\rho^2}-\log\big(1+\sqrt{\frac{\log n}{k}}\big)^2\Big).\nonumber
\end{align}
Since $\frac{\lVert\vtheta\rVert_2^2}{\rho^2}$ is generally large, the smallest $n$ is 2 and $\log\big(1+\sqrt{\frac{\log n}{k}}\big)^2 > 0$. Therefore, for $n>1$, $R_1'<0$, meaning $R_1$ decreases as $n$ increase.

\textbf{Analysis of $\boldsymbol{R_2}$:} The derivative for $R_2$ \wrt $n$ is 
\begin{equation}
    R_2'=\frac{4}{(n-1)^2}(1-\frac{1}{n}-\log n + \log \delta).\nonumber
\end{equation}
Since $\delta \leq 1$, for $n>1$, $R_2'<0$, indicating that $R_2$ decreases with increasing $n$.

\textbf{Analysis of $\boldsymbol{R_3}$:} The derivative for $R_3$ \wrt $n$ is
\begin{equation}
    R_3'=\frac{8\big(\frac{6(n-1)}{6n+3k}-\log(6n+3k)\big)}{(n-1)^2} < \frac{8\big(1 - \log(6n+3k)\big)}{(n-1)^2}.\nonumber
\end{equation}
For $n>1$, $\log(6n+3k)>1$, implying that $R_3'<0$ and $R_3$ decrease as $n$ increases.

\textbf{Conclusion.} For $n>1$, all terms $R_1$, $R_2$ and $R_3$ decreases as $n$ increases. Thus $R(\frac{\lVert\vtheta\rVert_2^2}{\rho^2}, \frac{1}{n})$ is a decreasing function of $n$. 

\section{Efficient \our variants}
\label{sup:sec:efficient-pace}
Building upon strong theoretical foundation of \our for generalization, we demonstrate that simple modifications can reduce  memory and training time requirements of PACE. In this section, we explore two efficient variants, \pacefast and \pacehalflazy, both maintaining similar computational and memory requirements as the baseline while improving performance. We then provide empirical results which show that \pacefast slightly outperforms \pacehalflazy while requiring no additional hyperparameters and using fewer computational resources. Given its superior efficiency, we further explore the potential of \pacefast for resource-efficient fine-tuning. By simply reducing the batch size and epochs, \pacefast outperforms the baseline while using significantly less GPU memory and training time.

\textbf{\pacefast}: Building on the observation that only small datasets are typically available for fine-tuning, we assume that the model behavior changes gradually across epochs. Under this assumption, we store the model outputs from the previous epoch ($f_{e-1}(\vx)$), which contain inherent noise due to the adapter perturbation, and compute the consistency regularization loss between these stored outputs and the current epoch's noised outputs:
\begin{equation}
    \dpace_\text{fast} (\vx) = \|f(\vx)-\vo_{e-1}\|_2^2; \quad \text{where}\quad \vo_{e-1}=f_{e-1}(\vx).
\end{equation}
Here the output vector $\vo\in\bbR^C$, where $C$ is the number of classes. Since $f(\cdot)$ applies noise perturbation to the adapter and changes gradually between epochs, $f_{e-1}(\vx)$ and $f(\vx)$ can be seen as applying different {\em\iid} noises to similar model states. This approach preserves the theoretical foundation of PACE while incurring minimal storage and computation costs. With typically few classes $C$ and a  limited number of samples in fine-tuning, storing $\vo_{e-1}$ within GPU or CPU memory is manageable.

\textbf{\pacehalflazy}: During training, the network always applies noise perturbations. Every $N$-th iteration uses a half batch size and consistency regularization, while all other iterations use the full batch size.

\textbf{Memory and computational efficiency of two variants.}
Both variants maintain similar computational and memory requirements as the baseline. To demonstrate this, we conduct experiments on CIFAR-100 (VTAB-1K) using ViT-B/16, Camelyon (VTAB-1K) with Swin-B, and ImageNet (domain adaptation) with ViT-B/16. Table \ref{tab:efficient_pace} compares maximum GPU memory usage, total training time, and accuracy for each task, showing that \pacefast and \pacehalflazy significantly improve upon the baseline while maintaining similar computational demands. 

We find that \pacefast slightly outperforms \pacehalflazy without requiring additional hyperparameters, yet it needs to store outputs from the previous epoch. We therefore analyze its memory requirements.

\begin{table}[h]
\fontsize{8}{9}\selectfont
\setlength{\tabcolsep}{0.07cm}
\renewcommand{\arraystretch}{1.3}
\centering
\caption{GPU memory usage, training time, and accuracy for \pacefast and \pacehalflazy. here, `m' denotes minutes, Both variants outperform the baseline while maintaining similar computational demands.}
\label{tab:efficient_pace}
\vspace{0.1cm}
\begin{tabular}{!{\vrule width \boldlinewidth}l|ccc|ccc|ccc!{\vrule width \boldlinewidth}}
\topline
\multirow{2}{*}{Method} & \multicolumn{3}{c|}{CIFAR-100 (ViT/16-B)}	& 
\multicolumn{3}{c|}{Camelyon (Swin-B)} & \multicolumn{3}{c!{\vrule width \boldlinewidth}}{ImageNet (ViT/16-B)}\\
\cline{2-10}
& GPU Memory & Time & Accuracy & GPU Memory & Time & Accuracy & GPU Memory &	Time &	Mean Acc. \\
\middleline
\baseline       & 8.9GB	& 29m &	74.6 &	15.7GB & 33m & 86.7 & 8.9GB &	161m &	44.8\\
\hline
\ +\our         & 17.7GB & 53m & 79.0 &	29.4GB &	60m & 89.3 & 17.7GB & 278m &	46.3\\
\ +\pacefast    & \textbf{9.0GB}  & \textbf{29m} & \textbf{78.3} &	15.7GB &	34m	& 88.8 & \textbf{9.0GB}  & \textbf{162m} & \textbf{46.1} \\
\ +\pacehalflazy ($N\!=\!2$) & 9.3GB	& 29m &	78.7 &	\textbf{15.7GB} & \textbf{36m} & \textbf{89.2} & 9.0GB & 165m & 46.0\\
\ +\pacehalflazy ($N\!=\!4$) & 9.3GB	& 29m &	78.4 &	15.7GB &	35m	& 88.9 & 9.0GB &	163m & 45.6\\
\ +\pacehalflazy ($N\!=\!6$) & 9.3GB	& 29m & 78.4 &	15.7GB &	35m & 89.0 & 9.0GB & 163m & 45.7\\
\ +\pacehalflazy ($N\!=\!10$) & 9.3GB	& 29m & 78.2 &	15.7GB &	35m	& 88.9 & 9.0GB &	162m & 45.6\\
\bottomline
\end{tabular}
\end{table}

\textbf{Memory efficiency of \pacefast.} We compare the additional memory requirement of \pacefast with the baseline GPU memory consumption. Table \ref{tab:mem_fast} shows that the memory overhead of \pacefast is negligible compared to the baseline GPU memory requirements and can be easily stored in GPU. Moreover, even in the rare scenario of fine-tuning on the full ImageNet 1K dataset (1.2 million samples), \pacefast requires only 4.8GB of additional memory for storing the output of the model's classification head. This is significantly smaller than the dataset itself (>100GB) and can be easily accommodated in the CPU/GPU memory.

\begin{table}[h]
\fontsize{8}{9}\selectfont
\setlength{\tabcolsep}{0.31cm}
    \centering
    \caption{Comparison of \pacefast memory overhead and the baseline GPU memory requirements.}
    \vspace{0.1cm}
    \label{tab:mem_fast}
    \begin{tabular}{!{\vrule width \boldlinewidth}l|c|c|c!{\vrule width \boldlinewidth}}
    \topline
    Dataset	& Memory of \pacefast & Baseline GPU Memory	& Ratio\\
    \middleline
    CIFAR-100 (VTAB-1K w/ ViT/16-B) &	390KB &	8.9GB &	0.0042\%\\
    Camelyon (VTAB-1K w/ Swin-B)	& 7.81KB & 15.7GB & 0.000047\%\\
    ImageNet (Domain adaptation w/ ViT/16-B) & 61MB &	8.9GB & 0.67\% \\
    \bottomline
    \end{tabular}
\end{table}

\textbf{Resource-Efficient training with \pacefast.} Given the superior performance, minimal memory overhead, and no need for additional hyperparameters of \pacefast, we explore its potential for resource-efficient training by maintaining the same number of updates with reduced batch size and proportionally reduced epochs. Table \ref{tab:resource_efficient} shows that even with 1/8  batch size and epochs, \pacefast still outperforms the baseline by 1.7\% while only using $\sim$1/3 GPU memory and $\sim$1/4 training time. This demonstrates the robustness and generalization benefits that \pacefast brings to models, enabling them to excel under constrained training configurations. Such an efficiency is particularly valuable for fine-tuning large foundation models, where resource constraints necessitate small batch sizes and typically lead to sharp loss landscapes, yet the theoretical guarantee of PACE for smooth loss landscapes provides a promising solution for these challenges.

\begin{table}[h]
\vspace{-0.2cm}
\fontsize{8}{9}\selectfont
\setlength{\tabcolsep}{0.058cm}
\renewcommand{\arraystretch}{1.37}
    \centering
    \caption{Results of \pacefast with a reduced batch size and epochs on CIFAR-100 (VTAB-1K w/ ViT-B/16), Camelyon (VTAB-1K w/ Swin-B), ImageNet (Domain adaptaion w/ ViT-B/16). \pacefast outperforms baseline while using less GPU memory and training time.}
    \label{tab:resource_efficient}
    \vspace{0.1cm}
    \begin{tabular}{!{\vrule width \boldlinewidth}l|ccc|ccc|ccc|ccc!{\vrule width \boldlinewidth}}
    \topline
    \multirow{2}{*}{Method} & \multicolumn{3}{c|}{CIFAR-100}	& \multicolumn{3}{c|}{Camelyon} & \multicolumn{3}{c|}{ImageNet} & \multicolumn{3}{c!{\vrule width \boldlinewidth}}{Average}\\
    \cline{2-13}
    & Mem. & Time & Acc. & Mem. & Time & Acc. & Mem. & Time & MeanAcc. & Mem. & Time & Acc.\\
    \middleline
    \baseline & 8.9GB & 29m & 74.6 & 15.7GB & 33m & 86.7 &	8.9GB & 161m & 44.8 & 11.1GB & 74m & 68.7\\
    \hline
    \ +\pacefast ($\frac{1}{2}\!$  batch size, $\frac{1}{2}\!$ epochs) & 5.4GB & 17m & 78.1 & 8.6GB & 21m & 88.9 & 5.4GB & 85m & 45.8 & 6.5GB & 41m & 70.9\\
    \ +\pacefast ($\frac{1}{4}\!$   batch size, $\frac{1}{4}\!$  epochs) & 3.5GB & 10m & 77.8 & 6.0GB & 14m & 88.7	& 3.5GB & 50m & 45.6 & 4.3GB & 25m & 70.7\\
    \ +\pacefast ($\frac{1}{8}\!$   batch size, $\frac{1}{8}\!$  epochs) & 2.9GB & 6m	& 77.2 & 5.2GB &	10m	& 88.6 & 2.9GB &	32m & 45.5 &	3.7GB & 16m & 70.4\\
    \bottomline
    \end{tabular}
\end{table}

\begin{table}[h]
\vspace{-0.2cm}
\fontsize{8}{9}\selectfont
\setlength{\tabcolsep}{0.5cm}
    \centering
    \caption{Classification results for different methods on VTAB-1K with different training epochs.}
    \label{tab:vtab_100epoch}
    \vspace{0.1cm}
    \begin{tabular}{!{\vrule width \boldlinewidth}l|c|ccc|c!{\vrule width \boldlinewidth}}
    \topline
    \#Epoch & Method & Natural &	Specialized & Structured & Avg.\\
    \middleline
    530	   & GLoRA	& 83.61	& 87.02	& 63.27 & 77.97\\
    \hline
    100	   & Baseline &	81.94 &	85.40 &	61.40 &	76.24\\
    100	   & +\our	& 83.94	& 87.44	& 64.62	& 78.67\\
    50	   & +\our (half batch size) & 83.77 &	87.32 &	63.92 &	78.34\\
    \hline
    200	   & Baseline & 82.28 &	85.30 &	61.64 &	76.40\\
    200	   & \ +\our	  & 84.13 &	87.57 &	64.85 &	78.85\\
    \hline
    300	   & Baseline &	82.41 &	85.00 &	61.80 &	76.40\\
    300	   & +\our	& 84.32	& 87.55	& 65.13 & 79.00\\
    \bottomline
    \end{tabular}
\end{table}

\section{Additional Experiments}
In this section, we provide additional experiments of \our on VTAB-1K with different epochs, varying training data sizes on FGVC benchmarks, self-supervised pre-trained backbones and combinations with other PEFT methods.

\subsection{Experiments of VTAB-1K with different epochs}
\label{sup:subsec:100epochs}
In Table \ref{tab:vtab-comparison}, We use 300 epochs for VTAB-1K tasks as we observed slight improvements over 100 epochs. However, this does not mean \our requires longer training to converge. Since the optimizer uses the cosine learning rate decay, reducing the number of training epochs to 100 has a minimal impact on performance, as shown in Table \ref{tab:vtab_100epoch}.

To ensure fair memory and computational budgets, we also tested \our with half the batch size and 50 epochs. Table \ref{tab:vtab_100epoch} shows that under these conditions, \our still improves baseline accuracy by 2.10\%, and outperforms the previous SOTA GLoRA, which uses 500 epochs for training and 30 for parameter search. These results demonstrate \our's efficiency and effectiveness across various training configurations.

\subsection{Experiments on FGVC with limited training data}
\label{sup:subsec:fgvc}
To validate generalization benefits of \our on limited data settings, we conduct experiments on FGVC using 50\%, 20\%, and 10\% of the original training samples. Table \ref{tab:fgvc_few} shows that PACE achieves larger improvements with smaller data sizes, aligning with our theoretical analyses.

\begin{table}[h]
\vspace{-0.2cm}
\fontsize{8}{9}\selectfont
\setlength{\tabcolsep}{0.15cm}
    \centering
    \caption{Classification results on FGVC using varying percentages of data based on ViT-B/16.}
    \label{tab:fgvc_few}
    \begin{tabular}{!{\vrule width \boldlinewidth}l|ccc|ccc|ccc|ccc|ccc!{\vrule width \boldlinewidth}}
    \topline
    \multirow{2}{*}{Method}	& \multicolumn{3}{c|}{CUB} & \multicolumn{3}{c|}{NAB} &\multicolumn{3}{c|}{Flowers} & \multicolumn{3}{c|}{Stanford Dogs} & \multicolumn{3}{c!{\vrule width \boldlinewidth}}{Stanford Cars}\\
    \cline{2-16}
    & 50\% & 20\% & 10\% & 50\% & 20\% & 10\% & 50\% & 20\% & 10\% & 50\% & 20\% & 10\%	& 50\% & 20\% & 10\%\\
    \middleline
    baseline & 87.1 & 83.9 & 79.1 & 80.7 & 75.0 & 70.2 & 98.5 & 96.5 & 93.1 & 90.6 & 88.7 &	86.9 & 78.7 & 54.9 &	30.1\\
     \rowcolor{tblcolor}\ +\our	 & 88.4 & 85.5 & 81.4 & 82.9 & 77.5 & 73.8 & 99.2 & 97.9 & 96.1 & 91.8 & 90.9 &	89.8 & 80.5	 & 57.3 & 33.2\\
    \bottomline
    \end{tabular}
\end{table}

\subsection{Experiments on self-supervised pre-trained backbones}
\label{sup:subsec:self-supervised}
To further verify the effectiveness of \our on a self-supervised pre-trained backbone, we conduct VTAB-1K experiments on SVHN, Camelyon, and Clevr-Count using MAE \cite{he2022masked} and DINO \cite{he2022masked}, with ViT-B/16 pre-trained on ImageNet-1K \cite{imagenet}. Table \ref{tab:self-supervised} shows that \our improves the baseline on these self-supervised backbones, confirming its applicability to fine-tuning self-supervised models.

\begin{table}[h]
\fontsize{8}{9}\selectfont
\setlength{\tabcolsep}{0.37cm}
    \centering
    \caption{Classification results on VTAB-1K using self-supervised DINO and MAE, with ViT-B/16 pre-trained on the ImageNet-1K dataset. }
    \label{tab:self-supervised}
    \vspace{0.1cm}
    \begin{tabular}{!{\vrule width \boldlinewidth}l|ccc|ccc!{\vrule width \boldlinewidth}}
    \topline
    \multirow{2}{*}{Method} & \multicolumn{3}{c|}{MAE} &	\multicolumn{3}{c!{\vrule width \boldlinewidth}}{DINO}\\
    \cline{2-7}
         & SVHN &	Camelyon &	Clevr-Count &	SVHN &	Camelyon &	Clevr-Count\\
    \middleline
    Full &  90.1 	 &      74.6  &        52.5	 &        89.7 &     73.1 &    34.5\\
    Linear&	44.5     &      79.9  &        57.1	 &        50.7 &	 82.5 &    44.2\\
    \baseline & 89.3 &  	82.7  &        82.1	 &        90.0 &	 85.4 &    55.7\\
     \rowcolor{tblcolor}\ +\our &	\textbf{93.5} &\textbf{85.8}  & \textbf{86.4}  &\textbf{91.7} & \textbf{88.1} &    \textbf{61.0}\\
    \bottomline
    \end{tabular}
\end{table}

\subsection{Experiments of Combining \our with Other PEFT}
\label{sup:subsec:other-peft}
We conducted experiments combining \our with several PEFT methods, including AdaptFormer \cite{chen2022adaptformer}, GLoRA \cite{glora}, COFT \cite{Qiu2023OFT}, and BOFT \cite{liu2024boft}, on CIFAR-100 (VTAB-1K) and ImageNet (domain adaptation) using ViT-B/16. Table \ref{tab:other_peft} shows that integrating \our improves the baseline performance.
\begin{table}[h]
\fontsize{8}{9}\selectfont
\setlength{\tabcolsep}{0.35cm}
    \centering
    \caption{Classification results of different PEFT methods based on ViT-B/16.}
    \label{tab:other_peft}
    \begin{tabular}{!{\vrule width \boldlinewidth}l|c|ccccc|c!{\vrule width \boldlinewidth}}
    \topline
    \multirow{2}{*}{Method}	& \multirow{2}{*}{CIFAR-100 (VTAB-1K)} & \multicolumn{6}{c!{\vrule width \boldlinewidth}}{ImageNet  (Domain Adaptation)}\\
    \cline{3-8}
                &  & Source	& -Sketch	& -V2	& -A &	-R & Avg.\\
    \middleline
    AdaptFormer & 70.6 & 77.4 & 26.5 & 67.4	& 12.4 & 28.7 &	42.4 \\
    \rowcolor{tblcolor} \ +\our 	& \textbf{74.8} & \textbf{78.2} &\textbf{27.4} & \textbf{67.9}	& \textbf{13.9} & \textbf{31.7} & \textbf{43.8} \\
    \hline
    GLoRA	    & 75.9 & 78.2 &	30.3 & 68.1	& 13.5 & 31.6 &	44.3 \\
     \rowcolor{tblcolor}\ +\our 	& \textbf{78.6} & \textbf{78.8} & \textbf{31.7} & \textbf{69.0} & \textbf{15.9} & \textbf{34.4} & \textbf{45.9} \\
    \hline
    COFT	   & 71.8 & 76.9 & 26.4	& 66.7	& 13.1	& 30.7	& 42.7\\
     \rowcolor{tblcolor}\ +\our	   & \textbf{75.3} & \textbf{77.8} & \textbf{27.9} & \textbf{68.2} & \textbf{14.9} & \textbf{32.9} & \textbf{44.3}\\
    \hline
    BOFT	   & 72.3 & 77.1 & 27.0	& 66.8	& 12.8	& 31.1	& 42.9\\
    \rowcolor{tblcolor} \ +\our	   & \textbf{75.7} & \textbf{77.9} & \textbf{28.3}	& \textbf{68.2}	& \textbf{14.7}	& \textbf{33.4}	& \textbf{44.5}\\
    \bottomline
    \end{tabular}
\end{table}

\section{Additional Plots}
Figures \ref{sup:fig:gradient_norm_training} and \ref{sup:fig:gn_lbds_swin} show the gradient issues in FPA and the gradient regularization effects of PACE.

\begin{figure}[h!]
    \centering
    \includegraphics[width=0.8\linewidth]{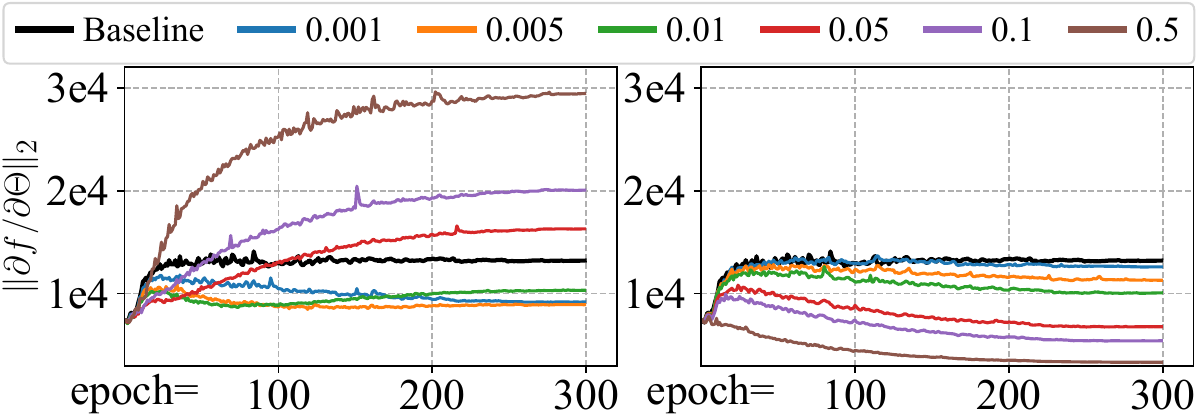}
    \begin{subfigure}{0.4\linewidth}
        \caption{FPA}\label{sup:subfig:gn_fpa}
    \end{subfigure}
    \begin{subfigure}{0.4\linewidth}
        \caption{PACE}\label{sup:subfig:gn_pace}
    \end{subfigure}
    \caption{Gradient norms of (a) FPA and (b) PACE with different regularization strengths $\lambda$ during training on CIFAR-100 (VTAB-1K) w/ ViT-B/16. Figure \ref{fig:analysis_gn} illustrates the average gradient norm over training epochs.}
    \label{sup:fig:gradient_norm_training}
\end{figure}

\begin{figure}[h!]
    \centering
    \includegraphics[width=0.8\linewidth]{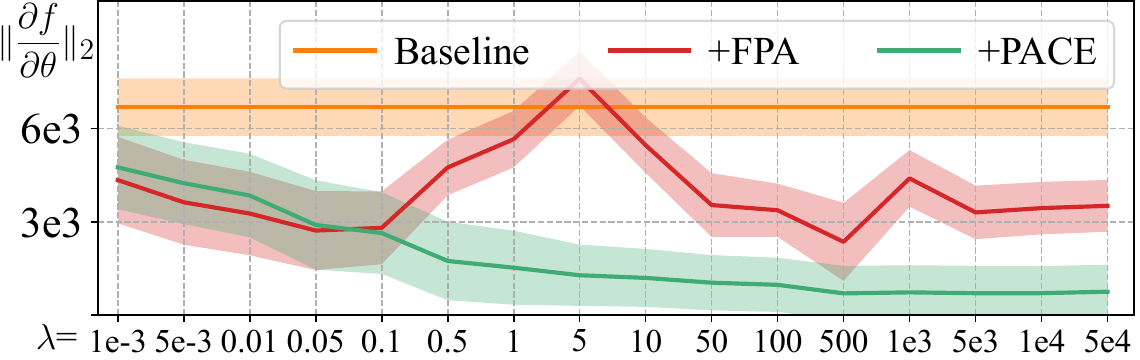}
    \caption{Gradient norms of models across wide range of regularization strengths $\lambda$ on Camelyon (VTAB-1K) w/ Swin-B. Line and shadow represent mean and std over training epochs. While gradient explosion is less frequent for FPA in this setting, it exhibits unpredictable gradient norm with varied regularization strengths. In contrast, \our reliably lowers gradient norms as regularization strength $\lambda$ increases, demonstrating its robustness for effective gradient control.}
    \label{sup:fig:gn_lbds_swin}
\end{figure}

\section{Hyperparameter settings}
For each dataset, we follow strategies from previous works \cite{Lian_2022_SSF, vpt, glora, repadapter} to apply grid search on the rank, learning rate and weight decay to establish strong baselines. Table \ref{sup:tab:vtab-hyper}, \ref{sup:tab:fewshot-hyper}, \ref{sup:tab:fgvc-hyper} and \ref{sup:tab:dg-hyper} present the hyperparameters and number of trainable parameters used in our strong baseline for VTAB-1K, few-shot learning, FGVC and domain adaptation tasks.

With these strong baselines, we apply grid search on $\lambda\in\{0.02, 0.05, 0.1, 0.2, 0.5, 1\}$ and $\sigma\in\{0.1, 0.5, 1, 1.5, 2\}$ for \our to optimize its performance.

\label{supp:baseline}
\begin{table}[h!]
\vspace{-0.2cm}
\begin{center}
\fontsize{8}{9}\selectfont
\setlength{\tabcolsep}{0.07cm}
\caption{Hyperparameters for baseline on VTAB-1K with ViT-B/16. A: \baseline, B: \loraadd. lr: learning rate. WD: weight decay.}
\label{sup:tab:vtab-hyper}
\begin{tabular}{!{\vrule width \boldlinewidth}l|ccccccc|cccc|cccccccc|c!{\vrule width \boldlinewidth}}
\topline
&\multicolumn{7}{c|}{Natural} & \multicolumn{4}{c|}{Specialized} & \multicolumn{8}{c|}{Structured} &\\
\cline{2-20}
\rotatebox{90}{Hyperparameter} &\rotatebox{90}{Cifar100} & \rotatebox{90}{Caltech101} & \rotatebox{90}{DTD} & \rotatebox{90}{Flowers102} & \rotatebox{90}{Pets} & \rotatebox{90}{SVHN} & \rotatebox{90}{Sun397} & \rotatebox{90}{Camelyon} & \rotatebox{90}{EuroSAT} & \rotatebox{90}{Resisc45} & \rotatebox{90}{Retinopathy} & \rotatebox{90}{Clevr-Count} & \rotatebox{90}{Clevr-Dist} & \rotatebox{90}{DMLab} & \rotatebox{90}{KITTI-Dist} & \rotatebox{90}{dSpr-Loc} & \rotatebox{90}{dSpr-Ori} & \rotatebox{90}{sNORB-Azim\ } & \rotatebox{90}{NsORB-Ele} & \rotatebox{90}{Average parameter (M)} \\
\middleline
Method & A & A & A & A & A & A & A & A & A & A & B & B & B & A & A & A & A & A & B & \multirow{4}{*}{1.81}\\
Rank & 10 & 14 & 12 & 18 & 18 & 14 & 10 & 8 & 8 & 10 & 2 & 2 & 8 & 18 & 4 & 10 & 10 & 22 & 4 & \\
lr & 1e-3 & 1e-3 & 1e-3 & 1e-3 & 1e-3 & 1e-2 & 1e-3 & 5e-3 & 5e-3 & 5e-3 & 5e-4 & 5e-4 & 1e-4 & 5e-3 & 5e-3 & 5e-3 & 5e-3 & 1e-2 & 2e-4 & \\
WD & 1e-4 & 1e-4 & 1e-3 & 1e-2 & 1e-3 & 1e-3 & 1e-2 & 1e-2 & 1e-2 & 1e-2 & 1e-4 & 1e-3 & 1e-4 & 1e-3 & 1e-3 & 1e-4 & 1e-2 & 1e-2 & 1e-2 & \\
\bottomline
\end{tabular}
\end{center}
\end{table}

\begin{table}[h!]
\vspace{-0.2cm}
\begin{center}
\fontsize{8}{9}\selectfont
\setlength{\tabcolsep}{0.19cm}
\caption{Ranks for baselines in Few-shot learning. Weight decay is fixed at 1e-4.}
\label{sup:tab:fewshot-hyper}
\begin{tabular}{!{\vrule width \boldlinewidth}l|ccccc|c!{\vrule width \boldlinewidth}}
\topline
\multirow{2}{*}{\diagbox{Baseline}{learning rate}} & FGVCAircraft & Food101 & Flowers102 & OxfordPets & StanfordCars & Mean \\
\cline{2-6}
& 5e-3 & 5e-3 & 5e-3 & 2e-3 & 2e-3 & Parameter (M)\\
\hline
\loraadd & 4 & 4 & 4 & 4 & 10 & 0.93 \\
\vptadd & 1 & 1 & 1 & 1 & 1 & 0.14 \\
\baseline & 14 & 10 & 18 & 18 & 24 & 2.70\\
\bottomline
\end{tabular}
\end{center}
\end{table}

\begin{table}[h!]
\vspace{-0.2cm}
\begin{center}
\fontsize{8}{9}\selectfont
\setlength{\tabcolsep}{0.18cm}
\caption{Hyperparameters for the baseline \baseline in FGVC.}
\label{sup:tab:fgvc-hyper}
\begin{tabular}{!{\vrule width \boldlinewidth}l|ccccc|c!{\vrule width \boldlinewidth}}
\topline
Hyperparameter & CUB-200-2011 & NABirds & OxfordFlowers & StanfordDogs & StanfordCars  & Mean Parameter (M)\\
\hline
learning rate & 5e-3 & 5e-4 & 5e-3 & 5e-3 & 2e-4 & \multirow{3}{*}{2.80}\\
weight decay & 1e-2 & 1e-3 & 1e-3 & 1e-2 & 1e-3 & \\
rank & 14 & 18 & 18 & 24 & 14 & \\
\bottomline
\end{tabular}
\end{center}
\end{table}

\begin{table}[h!]
\vspace{-0.2cm}
\begin{center}
\fontsize{8}{9}\selectfont
\setlength{\tabcolsep}{0.39cm}
\caption{Hyperparameters for baseline \baseline in domain adaptation.}
\label{sup:tab:dg-hyper}
\begin{tabular}{!{\vrule width \boldlinewidth}lcccc!{\vrule width \boldlinewidth}}
\topline
Baseline & rank & learning rate & weight decay & Parameter (M)\\
\hline
\baseline & 10 & 5e-4 & 1e-2  & 2.39 \\
\bottomline
\end{tabular}
\end{center}
\end{table}

\section{Experiment details for GSM-8K}
\label{sup:sec:gsm8k}
We conduct experiments on text generation tasks by fine-tuning Phi-3-mini-4k-instruct \cite{abdin2024phi} on the GSM-8K \cite{cobbe2021training} dataset using causal language modeling. We use learning rate of 2e-6, batch size of 4, LoRA rank of 16, prompt ``Answer below question. First think step-by-step and then answer the final number:\textbackslash n\textbackslash n<Question>'' as instruction and fine-tune models on the training set and evaluated the performance on the test set.

\clearpage
\newpage
\section*{NeurIPS Paper Checklist}

\renewcommand{\labelenumi}{\arabic{enumi}.}

\begin{enumerate}

\item {\bf Claims}
    \item[] Question: Do the main claims made in the abstract and introduction accurately reflect the paper's contributions and scope?
    \item[] Answer: \answerYes{} 
    \item[] Justification: We theoretically and empirically verify the claims and contributions made in the abstract and introduction.

\item {\bf Limitations}
    \item[] Question: Does the paper discuss the limitations of the work performed by the authors?
    \item[] Answer: \answerYes{} 
    \item[] Justification: The limitations of our work are discussed in \S \ref{sup:sec:limitations}

\item {\bf Theory Assumptions and Proofs}
    \item[] Question: For each theoretical result, does the paper provide the full set of assumptions and a complete (and correct) proof?
    \item[] Answer: \answerYes{} 
    \item[] Justification: Complete proofs for each theorem are provided in \S \ref{sup:sec:proof}.

    \item {\bf Experimental Result Reproducibility}
    \item[] Question: Does the paper fully disclose all the information needed to reproduce the main experimental results of the paper to the extent that it affects the main claims and/or conclusions of the paper (regardless of whether the code and data are provided or not)?
    \item[] Answer: \answerYes{} 
    \item[] Justification: Training details and hyperparameter selection are presented in Sec. \ref{sec:experiments} and \S \ref{supp:baseline}, respectively.

\item {\bf Open access to data and code}
    \item[] Question: Does the paper provide open access to the data and code, with sufficient instructions to faithfully reproduce the main experimental results, as described in supplemental material?
    \item[] Answer: \answerYes{} 
    \item[] Justification: We will release our code.

\item {\bf Experimental Setting/Details}
    \item[] Question: Does the paper specify all the training and test details (e.g., data splits, hyperparameters, how they were chosen, type of optimizer, etc.) necessary to understand the results?
    \item[] Answer: \answerYes{} 
    \item[] Justification: Experimental settings and details are presented in Sec. \ref{sec:experiments}.

\item {\bf Experiment Statistical Significance}
    \item[] Question: Does the paper report error bars suitably and correctly defined or other appropriate information about the statistical significance of the experiments?
    \item[] Answer: \answerYes{} 
    \item[] Justification: All reported results are averaged over three random seeds.

\item {\bf Experiments Compute Resources}
    \item[] Question: For each experiment, does the paper provide sufficient information on the computer resources (type of compute workers, memory, time of execution) needed to reproduce the experiments?
    \item[] Answer: \answerYes{} 
    \item[] Justification: All experiments were conducted on a single NVIDIA H100 GPU with 96 GB memory, with each experiment completing within 8 hours.
    
\item {\bf Code Of Ethics}
    \item[] Question: Does the research conducted in the paper conform, in every respect, with the NeurIPS Code of Ethics \url{https://neurips.cc/public/EthicsGuidelines}?
    \item[] Answer: \answerYes{}, 
    \item[] Justification: We have carefully reviewed and adhered to the code of ethics throughout our research and writing process.

\item {\bf Broader Impacts}
    \item[] Question: Does the paper discuss both potential positive societal impacts and negative societal impacts of the work performed?
    \item[] Answer: \answerYes{} 
    \item[] Justification: Potential impacts are discussed in \S \ref{sup:sec:limitations}.
    
\item {\bf Safeguards}
    \item[] Question: Does the paper describe safeguards that have been put in place for responsible release of data or models that have a high risk for misuse (e.g., pre-trained language models, image generators, or scraped datasets)?
    \item[] Answer: \answerNA{}. 
    \item[] Justification: Our work poses no such risks.

\item {\bf Licenses for existing assets}
    \item[] Question: Are the creators or original owners of assets (e.g., code, data, models), used in the paper, properly credited and are the license and terms of use explicitly mentioned and properly respected?
    \item[] Answer: \answerYes{}  
    \item[] Justification: All publicly available assets (models, code, and data) used in this work have been properly credited, and their respective licenses and terms of use have been explicitly mentioned and adhered to.

\item {\bf New Assets}
    \item[] Question: Are new assets introduced in the paper well documented and is the documentation provided alongside the assets?
    \item[] Answer: \answerNA{} 
    \item[] Justification: We do not release new assets in the submission.

\item {\bf Crowdsourcing and Research with Human Subjects}
    \item[] Question: For crowdsourcing experiments and research with human subjects, does the paper include the full text of instructions given to participants and screenshots, if applicable, as well as details about compensation (if any)? 
    \item[] Answer: \answerNA{}  
    \item[] Justification: the paper does not involve crowdsourcing nor research with human subjects.

\item {\bf Institutional Review Board (IRB) Approvals or Equivalent for Research with Human Subjects}
    \item[] Question: Does the paper describe potential risks incurred by study participants, whether such risks were disclosed to the subjects, and whether Institutional Review Board (IRB) approvals (or an equivalent approval/review based on the requirements of your country or institution) were obtained?
    \item[] Answer: \answerNA{} 
    \item[] Justification: the paper does not involve crowdsourcing nor research with human subjects.

\end{enumerate}

\end{document}

%% file: preamble.tex
\usepackage{array}
\usepackage{multirow}
\usepackage{booktabs}
\usepackage{makecell} 
\usepackage{graphicx}
\usepackage{amsmath}
\usepackage{subcaption}
\usepackage{diagbox}
\usepackage{xspace}
\usepackage{mathrsfs}
\usepackage{enumitem}

\usepackage{utfsym} 
\newcommand{\heart}{$\;\!$\usym{2665}}

\makeatletter
\DeclareRobustCommand\onedot{\futurelet\@let@token\mathbfv@onedotaux}
\def\mathbfv@onedotaux{\ifx\@let@token.\else.\null\fi\xspace}
\def\eg{\emph{e.g}\onedot} 
\def\ie{\emph{i.e}\onedot}

\def\wrt{w.r.t\onedot}

\def\iid{i.i.d\onedot }
\def\etal{\emph{et al}\onedot}

\newlength{\boldlinewidth}
\newcommand{\topline}{\Xhline{1.3pt}}
\newcommand{\bottomline}{\Xhline{1.3pt}}
\newcommand{\middleline}{\Xhline{1pt}}

\newcommand{\loraadd}{LoRA$_\text{add}$\xspace}
\newcommand{\loramul}{LoRA$_\text{mul}$\xspace}
\newcommand{\baseline}{LoRA$_\text{mul}$+VPT$_\text{add}$\xspace}
\newcommand{\vptadd}{VPT$_\text{add}$}

\newcommand{\our}{PACE\xspace}
\newcommand{\pacefast}{PACE$_\text{fast}$\xspace}
\newcommand{\pacehalflazy}{PACE$_\text{lazy}^\text{half}$\xspace}

\newcommand{\bbR}{\mathbb{R}}
\newcommand{\vx}{\boldsymbol{x}}
\newcommand{\va}{\boldsymbol{a}}
\newcommand{\vo}{\boldsymbol{o}}
\newcommand{\vtheta}{\boldsymbol{\theta}}
\newcommand{\mW}{\boldsymbol{W}}
\newcommand{\vb}{\boldsymbol{b}}
\newcommand{\din}{d_\text{in}}
\newcommand{\dout}{d_\text{out}}
\newcommand{\Wdown}{\mW_{\!\text{d}}}
\newcommand{\Wup}{\mW_{\!\text{u}}}
\newcommand{\vblora}{\vb_\text{lora}}
\newcommand{\vzero}{\boldsymbol{0}}

\newcommand{\mP}{\boldsymbol{P}}

\newcommand{\calDn}{\mathcal{D}^n}
\newcommand{\calL}{\mathcal{L}}
\newcommand{\scrD}{\mathscr{D}}
\newcommand{\vy}{\boldsymbol{y}}
\newcommand{\vxi}{\vx_{i}}
\newcommand{\vyi}{\vy_{i}}
\newcommand{\vepsilon}{\boldsymbol{\epsilon}}
\newcommand{\nablatheta}{\boldsymbol{\nabla}_{\vtheta}}
\newcommand{\Htheta}{\boldsymbol{H}_{\vtheta}}
\newcommand{\lambdaHmax}{\lambda_\text{max}^{\boldsymbol{H}}}
\newcommand{\vv}{\boldsymbol{v}}
\newcommand{\vnabla}{\boldsymbol{\nabla}}
\newcommand{\mH}{\boldsymbol{H}}

\newcommand{\Dfp}{D^\text{fp}}
\newcommand{\Dpace}{D^\text{pace}}
\newcommand{\bbE}{\mathbb{E}}
\newcommand{\vz}{\boldsymbol{z}}
\newcommand{\calN}{\mathcal{N}}
\newcommand{\dpace}{d^\text{pace}}
\newcommand{\vm}{\boldsymbol{m}}
\newcommand{\vu}{\boldsymbol{u}}
\newcommand{\vone}{\boldsymbol{1}}
\newcommand{\mI}{\boldsymbol{I}}
\newcommand{\mX}{\boldsymbol{X}}
\newcommand{\mZ}{\boldsymbol{Z}}
\newcommand{\vgamma}{\boldsymbol{\gamma}}
\newcommand{\vbeta}{\boldsymbol{\beta}}

\newtheorem{theorem}{Theorem}
\newtheorem{prop}{Proposition}
\newtheorem{lemma}{Lemma}